\documentclass[format=manuscript,authorversion]{extarticle}
\usepackage{mathptmx}
\usepackage{helvet}
\usepackage{courier}
\usepackage[latin9]{inputenc}
\usepackage{geometry}
\usepackage{color}
\usepackage{array}
\usepackage{verbatim}
\usepackage{booktabs}
\usepackage{dvipost}
\usepackage{amsmath}
\usepackage{graphicx}
\usepackage[authoryear]{natbib}
\usepackage[unicode=true]
 {hyperref}
\usepackage{breakurl}

\makeatletter

\providecommand{\tabularnewline}{\\}
\dvipostlayout
\dvipost{osstart color push Red}
\dvipost{osend color pop}
\dvipost{cbstart color push Blue}
\dvipost{cbend color pop}

\newcommand{\lyxaddress}[1]{
	\par {\raggedright #1
	\vspace{1.4em}
	\noindent\par}
}

\usepackage{multicol}

\usepackage{fancyhdr}
\makeatother

\begin{document}

\title{Adversarial Machine Learning Attacks and Defense Methods in the Cyber
Security Domain}

\author{Ishai Rosenberg, Asaf Shabtai, Yuval Elovici, and Lior Rokach}
\maketitle

\lyxaddress{Ben-Gurion University of the Negev}
\begin{abstract}
In recent years machine learning algorithms, and more specifically
deep learning algorithms, have been widely used in many fields, including
cyber security. However, machine learning systems are vulnerable to
adversarial attacks, and this limits the application of machine learning,
especially in non-stationary, adversarial environments, such as the
cyber security domain, where actual adversaries (e.g., malware developers)
exist. This paper comprehensively summarizes the latest research on
adversarial attacks against security solutions based on machine learning
techniques and illuminates the risks they pose. First, the adversarial
attack methods are characterized based on their stage of occurrence,
and the attacker\textquoteright s goals and capabilities. Then, we
categorize the applications of adversarial attack and defense methods
in the cyber security domain. Finally, we highlight some characteristics
identified in recent research and discuss the impact of recent advancements
in other adversarial learning domains on future research directions
in the cyber security domain. This paper is the first to discuss the
unique challenges of implementing end-to-end adversarial attacks in
the cyber security domain, map them in a unified taxonomy, and use
the taxonomy to highlight future research directions.
\end{abstract}
CCSDescription: General and reference -\textgreater{} Surveys and
overviews\\


\section{Introduction}

The growing use of machine learning, and particularly deep learning,
in fields like computer vision and natural language processing (NLP)
has been accompanied by increased interest in the domain of adversarial
machine learning (a.k.a. adversarial learning), i.e., attacking and
defending machine learning models algorithmically (\citet{DBLP:conf/ccs/HuangJNRT11}).
Of special interest are adversarial examples, which are samples modified
in order to be misclassified by the classifier attacked. 

Most of the research in adversarial learning has focused on the computer
vision domain, and more specifically in the image recognition domain.
This research has centered mainly on convolutional neural networks
(CNNs), which are commonly used in the computer vision domain (\citet{DBLP:journals/access/AkhtarM18,Qiu19}).
However, in recent years, adversarial example generation methods have
increasingly been utilized in other domains, including natural language
processing (e.g., \citet{JiDeepWordBug18}). These attacks have also
been used recently in the cyber security domain (e.g., \citet{DBLP:conf/raid/RosenbergSRE18}).
This domain is particularly interesting, because it is rife with adversaries
(e.g., malware developers who want to evade machine and deep learning-based
next generation antivirus products, spam filters, etc.). Adversarial
learning methods have already been executed against static analysis
deep neural networks\footnote{\href{https://skylightcyber.com/2019/07/18/cylance-i-kill-you/}{https://skylightcyber.com/2019/07/18/cylance-i-kill-you/}}. 

The main goal of this paper is to illuminate the risks posed by adversarial
learning to cyber security solutions that are based on machine learning
techniques. This paper contains: (1) an in-depth discussion about
the unique challenges of adversarial learning in the cyber security
domain (Section \ref{sec:Preliminary-Discussion:-The}), (2) a summary
of the state-of-the-art adversarial learning research papers in the
cyber security domain, categorized by application (Sections \ref{sec:Cyber-Applications-of}
and \ref{sec:Adversarial-Defense-Methods}) and characterized by our
unified taxonomy (defined in Section \ref{sec:Taxonomy}), (3) a discussion
of the challenges associated with adversarial learning in the cyber
security domain and possible future research directions (Section \ref{sec:Adversarial-Learning-FutureUniqu}),
including issues relating to existing defense methods (and the lack
thereof), and (4) a summary of the theoretical background on the adversarial
methods used in the cyber security domain (Section \ref{sec:Adversarial-Learning-Methods}).
We focus on adversarial attacks and defenses against classifiers used
in the cyber security domain, and not on other topics, such as attacks
on models' interpretability (\citet{Kuppa20}) or methods to assist
the model's interpretability (\citet{Ross_Doshi-Velez_2018}).

The main contributions of this paper are as follows:

1. We focus on a wide range of adversarial learning applications in
the cyber security domain (e.g., malware detection, speaker recognition,
cyber-physical systems, etc.), introduce a new unified taxonomy and
illustrate how existing research fits into this taxonomy, providing
a holistic overview of the field. In contrast, previous work focused
mainly on specific domains, e.g., malware detection or network intrusion
detection.

2. Using our taxonomy, we highlight research gaps in the cyber security
domain that have already been addressed in other adversarial learning
domains (e.g., Trojan neural networks in the image recognition domain)
and discuss their impact on current and future trends in adversarial
learning in the cyber security domain.

3. We discuss the unique challenges that attackers and defenders face
in the cyber security domain, challenges which do not exist in other
domains (e.g., image recognition). For instance, in the cyber security
domain, the attacker must verify that the original functionality of
the modified malware remains intact. Our discussion addresses the
fundamental differences between adversarial attacks performed in the
cyber security domain and those performed in other domains.

\section{\label{sec:Preliminary-Discussion:-The}Preliminary Discussion: The
Differences Between Adversarial Attacks in the Computer Vision and
Cyber Security Domains}

Most adversarial attacks published, including those published at academic
cyber security conferences, have focused on the computer vision domain,
e.g., generating a cat image that would be classified as a dog by
the classifier. However, the cyber security domain (e.g., malware
detection) seems to be a more relevant domain for adversarial attacks,
because in the computer vision domain, there is no real adversary
(with a few exceptions, e.g., terrorists who want to tamper with autonomous
cars\textquoteright{} pedestrian detection systems (\citet{roadsigns17}),
deepfakes (\citet{Liu_2020_CVPR}) which might cause fake news or
financial fraud, etc.). In contrast, in the cyber security domain,
there are actual adversaries with clear targeted goals. Examples include
ransomware developers who depend on the ability of their ransomware
to evade anti-malware products that would prevent both its execution
and the developers from collecting the ransom money, and other types
of malware that need to steal user information (e.g., keyloggers),
spread across the network (worms), or perform any other malicious
functionality while remaining undetected. 

A key step in defining the proper taxonomy for the cyber security
domain is answering this question: Given the obvious relevance of
the cyber security domain to adversarial attacks, why do most adversarial
learning researchers focus on computer vision? In addition to the
fact that image recognition is a popular machine learning research
topic, another major reason for the focus on computer vision is that
performing an end-to-end adversarial attack in the cyber security
domain is more difficult than performing such an attack in the computer
vision domain. The differences between adversarial attacks performed
in those two domains and the challenges that arise in the cyber security
domain are discussed in the subsections that follow.

\subsection{\label{subsec:ExecutableMaintaining-(Malicious}Keeping (Malicious)
Functionality Intact in the Perturbed Sample}

Any adversarial executable file must preserve its malicious functionality
after the sample\textquoteright s modification. This might be the
main difference between the image classification and malware detection
domains, and pose the greatest challenge. In the image recognition
domain, the adversary can change every pixel\textquoteright s color
(to a different valid color) without creating an \textquotedblleft invalid
picture\textquotedblright{} as part of the attack. However, in the
cyber security domain, modifying an API call or arbitrary executable\textquoteright s
content byte value might cause the modified executable to perform
a different functionality (e.g., changing a \emph{WriteFile()} call
to \emph{ReadFile()} ) or even crash (if you change an arbitrary byte
in an opcode to an invalid opcode that would cause an exception).
The same is true for network packets; perturbing a network packet
in order to evade the network intrusion detection system (NIDS) while
keeping a valid packet structure is challenging.

In order to address this challenge, adversaries in the cyber security
domain must implement their own methods (which are usually feature-specific)
to modify features in a way that will not break the functionality
of the perturbed sample, whether it is an executable, a network packet,
or something else. For instance, the adversarial attack used in \citet{DBLP:conf/raid/RosenbergSRE18}
generates a new malware PE (portable executable) with a modified API
call trace in a functionality-preserving manner.

\subsection{\label{subsec:Small-Perturbations-Are}Small Perturbations Are Not
Applicable for Discrete Features}

In the computer vision domain, gradient-based adversarial attacks,
e.g., the fast gradient sign method (FGSM) (Section \ref{sec:Adversarial-Learning-Methods}),
generate a minimal random modification to the input image in the direction
that would most significantly impact the target classifier prediction.
A \textquoteleft small modification\textquoteright{} (a.k.a. perturbation)
can be, for example, changing a single pixel\textquoteright s color
to a very similar color (a single pixel\textquoteright s color can
be changed from brown to black to fool the image classifier).

However, the logic of a \textquoteleft small perturbation\textquoteright{}
cannot be applied to many cyber security features. Consider a dynamic
analysis classifier that uses API calls. An equivalent to changing
a single pixel\textquoteright s color would be to change a single
API call to another API call. Even if we disregard what such a modification
would do to the executable\textquoteright s functionality (mentioned
in the previous subsection), would one of the following be considered
a \textquoteleft small perturbation\textquoteright{} of the \emph{WriteFile()}
API call: (1) modifying it to \emph{ReadFile()} (a different operation
for the same medium), or (2) modifying it to \emph{RegSetValueEx()}
(a similar operation for a different medium)? The use of discrete
features (e.g., API calls) which are not continuous or ordinal severely
limits the use of gradient-based attack methods (Section \ref{sec:Adversarial-Learning-Methods}).
The implications of this issue will be discussed in Section \ref{subsec:Evaluating-The-Robustness}.

\subsection{\label{subsec:Executables-are-More}Executables are More Complex
than Images}

An image used as input to an image classifier (usually a convolutional
neural network, CNN) is represented as a fixed size matrix of pixel
colors. If the actual image has different dimensions than the input
matrix, the picture will usually be resized, clipped, or padded to
fit the dimension limits. 

An executable, on the other hand, has a variable length: executables
can range in size from several kilobytes to several gigabytes. It
is also unreasonable to expect a clipped executable to maintain its
original classification. Let\textquoteright s assume that we have
a 100MB benign executable into which we inject a shellcode in a function
near the end-of-file. If the shellcode is clipped in order to fit
the malware classifier\textquoteright s dimensions, there is no reason
that the file would be classified as malicious, because its benign
variant would be clipped to the exact same form.

In addition, the execution path of an executable may depend on the
input, and thus, the adversarial perturbation should support any possible
input that the malware may encounter when executed in the target machine.

Finally, in the cyber security domain, classifiers usually use more
than a single feature type as input (e.g., phishing detection using
both URLs and connected server properties as was done in \citet{Shirazi19}).
{]}). A non-exhaustive list of features used in the cyber security
domain is presented in Figure \ref{fig:Overview-of-Our}. Some feature
types are easier to modify without harming the executable's functionality
than others. For instance, in the adversarial attack used in \citet{DBLP:conf/raid/RosenbergSRE18},
appending printable strings to the end of a malware PE file is much
easier than adding API calls to the PE file using a dedicated framework
built for this purpose. In contrast, in an image adversarial attack,
modifying each pixel has the same level of difficulty. The implications
of this issue are discussed in Section \ref{subsec:Evaluating-The-Robustness}.

While this is a challenge for malware classifier implementation, it
also affects adversarial attacks against malware classifiers. For
instance, attacks in which you have a fixed input dimension, (e.g.,
a 28{*}28 matrix for MNIST images), are much easier to implement than
attacks for which you need to consider the variable file size.

\textcolor{red}{}%

\section{\label{sec:Adversarial-Learning-Methods}Adversarial Learning Methods
Used in the Cyber Security Domain}

This section includes a list of theoretical adversarial learning methods
that are used in the cyber security domain but were inspired by attacks
from other domains. Due to space limitations, this is \textbf{not}
a complete list of the state-of-the-art prior work in other domains,
such as image recognition or NLP. Only methods that have been used
in the cyber security domain are mentioned. A more comprehensive list
that includes methods from other domains can be found, e.g., in \citet{Qiu19}.

The search for adversarial examples as a similar minimization problem
is formalized in \citet{DBLP:journals/corr/SzegedyZSBEGF13} and \citet{Biggio2013}:

\begin{equation}
\arg_{\boldsymbol{r}}\min f(\boldsymbol{x}+\boldsymbol{r})\neq f(\boldsymbol{x})\:s.t.\:\boldsymbol{x}+\boldsymbol{r}\in\boldsymbol{D}\label{eq:}
\end{equation}
\textcolor{red}{}%

The input \textbf{$\boldsymbol{x}$}, correctly classified by the
classifier $f$, is perturbed with \textbf{$\boldsymbol{r}$} such
that the resulting adversarial example, \textbf{$\boldsymbol{x}+\boldsymbol{r}$},
remains in the input domain \textbf{$\boldsymbol{D}$} but is assigned
a different label than \textbf{$\boldsymbol{x}$}. To solve Equation
\ref{eq:}, we need to transform the constraint $f(\boldsymbol{x}+\boldsymbol{r})\neq f(\boldsymbol{x})$
into an optimizable formulation. Then we can easily use the Lagrange
multiplier to solve it. To do this, we define a loss function $Loss()$
to quantify this constraint. This loss function can be the same as
the training loss, or different, e.g., hinge loss or cross-entropy
loss.

\subsection*{Gradient-Based Attacks }

In \emph{gradient-based attacks}, adversarial perturbations are generated
in the direction of the gradient, i.e., in the direction with the
maximum effect on the classifier's output (e.g., FGSM; Equation \ref{eq:-1}).
Gradient-based attacks are effective but require adversarial knowledge
about the targeted classifier's gradients. Such attacks require knowledge
about the architecture of the target classifier and are therefore
white-box attacks.

When dealing with malware classification tasks, differentiating between
malicious ($f(\boldsymbol{x})=1$) and benign ($f(\boldsymbol{x})=-1$),
as done by SVM, \citet{Biggio2013} suggested solving Equation \ref{eq:}
using gradient ascent. To minimize the size of the perturbation and
maximize the adversarial effect, the white-box perturbation should
follow the gradient direction (i.e., the direction providing the greatest
increase in model value, from one label to another). Therefore, the
perturbation $\boldsymbol{r}$ in each iteration is calculated as:

\begin{equation}
\boldsymbol{r}=\epsilon\nabla_{\boldsymbol{x}}Loss_{f}(\boldsymbol{x}+\boldsymbol{r},-1)\:s.t.\:f(\boldsymbol{x})=1
\end{equation}
\\
where $\epsilon$ is a parameter controlling the magnitude of the
perturbation introduced. By varying $\epsilon$, this method can find
an adversarial sample \textbf{$\boldsymbol{x}+\boldsymbol{r}$}.

\citet{DBLP:journals/corr/SzegedyZSBEGF13} views the (white-box)
adversarial problem as a constrained optimization problem, i.e., find
a minimum perturbation in the restricted sample space. The perturbation
is obtained by using box-constrained L-BFGS to solve the following
equation:

\begin{equation}
\arg_{\boldsymbol{r}}\min(d*|\boldsymbol{r}|+Loss_{f}(\boldsymbol{x}+\boldsymbol{r},l))\:s.t.\:\boldsymbol{x}+\boldsymbol{r}\in\boldsymbol{D}
\end{equation}
\\
where $d$ is a term added for the Lagrange multiplier.

\citet{Goodfellow14} introduced the white-box FGSM attack. This method
optimizes over the $L_{\infty}$ norm (i.e., reduces the maximum perturbation
on any input feature) by taking a single step to each element of \textbf{$\boldsymbol{r}$}
in the direction opposite the gradient. The intuition behind this
attack is to linearize the cost function $Loss()$ used to train a
model $f$ around the neighborhood of the training point \textbf{$\boldsymbol{x}$}
with a label $l$ that the adversary wants to force the misclassification
of. Under this approximation:

\begin{equation}
\boldsymbol{r}=\epsilon sign(\nabla_{\boldsymbol{x}}Loss_{f}(\boldsymbol{x},l))\label{eq:-1}
\end{equation}

\citet{Kurakin2016AdversarialEI} extended this method with the iterative
gradient sign method (iGSM). As its name suggests, this method applies
FGSM iteratively and clips pixel values of intermediate results after
each step to ensure that they are close to the original image (the
initial adversarial example is the original input): 

\begin{equation}
\boldsymbol{x'}_{n+1}=Clip\left\{ \boldsymbol{x'}_{n}+\epsilon sign(\nabla_{\boldsymbol{x}}Loss_{f}(\boldsymbol{x'}_{n},l))\right\} 
\end{equation}

The white-box Jacobian-based saliency map approach (JSMA) was introduced
by \citet{Papernot2016}. This method minimizes the $L_{0}$ norm
by iteratively perturbing features of the input which have large adversarial
saliency scores. Intuitively, this score reflects the adversarial
goal of taking a sample away from its source class and moves it towards
a chosen target class.

First, the adversary computes the Jacobian of the model: $\left[\frac{\partial f_{j}}{\partial x_{i}}(\boldsymbol{x})\right]_{i,j}$
where component $(i,j)$ is the derivative of class $j$ with respect
to input feature $i$. To compute the adversarial saliency map, the
adversary then computes the following for each input feature $i$:

\begin{equation}
S(\boldsymbol{x},t)[i]=\begin{cases}
0\;\;\;\;\;if\:\frac{\partial f_{t}(\boldsymbol{x})}{\partial x_{i}}<0\:or\:\sideset{}{_{j\neq t}}\sum\frac{\partial f_{j}(\boldsymbol{x})}{\partial x_{i}}>0\\
\frac{\partial f_{t}(\boldsymbol{x})}{\partial x_{i}}|\sideset{}{_{j\neq t}}\sum\frac{\partial f_{j}(\boldsymbol{x})}{\partial x_{i}}|\;\;\;\;\;otherwise
\end{cases}
\end{equation}
\\
where $t$ is the target class that the adversary wants the machine
learning model to assign. The adversary then selects the input feature
$i$ with the highest saliency score $S(\boldsymbol{x},t)[i]$ and
increases its value. This process is repeated until misclassification
in the target class is achieved or the maximum number of perturbed
features has been reached. This attack creates smaller perturbations
with a higher computing cost than the attack presented in \citet{Goodfellow14}.

The Carlini-Wagner (C\&W) attack \citet{DBLP:conf/sp/Carlini017}
formulates the generation of adversarial examples as an optimization
problem: find some small change \textbf{$\boldsymbol{r}$} that can
be made to an input \textbf{$\boldsymbol{x}+\boldsymbol{r}$} that
will change its classification, such that the result is still in the
valid range. They instantiate the distance metric with an \textbf{$L_{p}$}
norm (e.g., can either minimize the $L_{2}$, $L_{0}$ or $L_{1}$
distance metric), define the cost function $Loss()$ such that $Loss(\boldsymbol{x}+\boldsymbol{r})\geq0$
if and only if the model correctly classifies \textbf{$\boldsymbol{x}+\boldsymbol{r}$}
(i.e., gives it the same label that it gives \textbf{$\boldsymbol{x}$}),
and minimize the sum with a trade-off constant \textbf{$c$} which
is chosen by modified binary search:

\begin{equation}
\arg_{\boldsymbol{r}}\min\left(||\boldsymbol{r}||{}_{p}+c*Loss_{f}(\boldsymbol{x}+\boldsymbol{r},t)\right)\:s.t.\:\boldsymbol{x}+\boldsymbol{r}\in\boldsymbol{D}
\end{equation}
\\
where the cost function $Loss()$ maximizes the difference between
the target class probability and the class with the highest probability.
It is defined as:

\begin{equation}
\max\left(\arg_{i\neq t}\max(f(\boldsymbol{x}+\boldsymbol{r},i))-f(\boldsymbol{x}+\boldsymbol{r},t),-k\right)\label{eq:-2}
\end{equation}
\\
where $k$ is a constant to control the confidence.

\citet{DBLP:conf/cvpr/Moosavi-Dezfooli16} proposed the DeepFool adversarial
method to find the closest distance from the original input to the
decision boundary of adversarial examples. DeepFool is an untargeted
attack technique optimized for the $L_{2}$ distance metric. An iterative
attack by linear approximation is proposed in order to overcome the
nonlinearity at a high dimension. If $f$ is a binary differentiable
classier, an iterative method is used to approximate the perturbation
by considering that $f$ is linearized around \textbf{$\boldsymbol{x}+\boldsymbol{r}$}
in each iteration. The minimal perturbation is provided by:

\begin{equation}
\arg_{\boldsymbol{r}}\min\left(||\boldsymbol{r}||{}_{2}\right)\:s.t.\:f(\boldsymbol{x}+\boldsymbol{r})+\nabla_{\boldsymbol{x}}f(\boldsymbol{x}+\boldsymbol{r})^{T}*r=0
\end{equation}
\\
This result can be extended to a more general $L_{p}$ norm, $p\in[0,\infty)$.

\citet{DBLP:conf/iclr/MadryMSTV18} proposed a projected gradient
descent (PGD) based adversarial method to generate adversarial examples
with minimized empirical risk and the trade-off of a high perturbation
cost. The model\textquoteright s empirical risk minimization (ERM)
is defined as $E(x,y)_{\sim D}[Loss(x,y,\theta)]$, where $\boldsymbol{x}$
is the original sample, and y is the original label. By modifying
the ERM definition by allowing the adversary to perturb the input
$\boldsymbol{x}$ by the scalar value S, ERM is represented by $\min_{\theta}\rho(\theta):\rho(\theta)=E(x,y)_{\sim D}[max_{\delta\in S}Loss(\boldsymbol{x}+\boldsymbol{r},y,\theta)]$,
where $\rho(\theta)$ denotes the objective function. Note that \textbf{$\boldsymbol{x}+\boldsymbol{r}$}
is updated in each iteration.

\citet{DBLP:conf/aaai/ChenSZYH18} presented the elastic net adversarial
method (ENM). This method limits the total absolute perturbation across
the input space, i.e., the $L_{1}$ norm. ENM produces the adversarial
examples by expanding an iterative $L_{2}$ attack with an $L_{1}$
regularizer.

\citet{Papernot2016a} presented a white-box adversarial example attack
against RNNs. The adversary iterates over the words \textbf{$\boldsymbol{x}[i]$}
in the review and modifies it as follows:

\begin{equation}
\boldsymbol{x}[i]=\arg\min_{\boldsymbol{z}}||sign(\boldsymbol{x}[i]-\boldsymbol{z})-sign(J_{f}(\boldsymbol{x})[i,f(\boldsymbol{x})])||\:s.t.\:\boldsymbol{z}\in\boldsymbol{D}
\end{equation}
\\
where $f(\boldsymbol{x})$ is the original model label for \textbf{$\boldsymbol{x}$},
and $J_{f}(\boldsymbol{x})[i,j]=\frac{\partial f_{j}}{\partial x_{i}}(\boldsymbol{x})$.
$sign(J_{f}(\boldsymbol{x})[i,f(\boldsymbol{x})])$ provides the direction
one has to perturb each of the word embedding components in order
to reduce the probability assigned to the current class and thus change
the class assigned to the sentence. However, the set of legitimate
word embeddings is finite. Thus, one cannot set the word embedding
coordinates to any real value. Instead, one finds the word \textbf{$\boldsymbol{z}$}
in dictionary \textbf{$\boldsymbol{D}$} such that the sign of the
difference between the embeddings of \textbf{$\boldsymbol{z}$} and
the original input word is closest to $sign(J_{f}(\boldsymbol{x})[i,f(\boldsymbol{x})])$.
This embedding considers the direction closest to the one indicated
by the Jacobian as most impactful on the model\textquoteright s prediction.
By iteratively applying this heuristic to a word sequence, an adversarial
input sequence misclassified by the model is eventually found. 

\subsection*{Score-Based Attacks}

\emph{Score-based attacks} are based on knowledge of the target classifier's
confidence score. Therefore, these are gray-box attacks.

The zeroth-order optimization (ZOO) attack was presented in \citet{Chen2017}.
ZOO uses hinge loss in Equation \ref{eq:-2}:

\begin{equation}
\max\left(\arg_{i\neq t}\max(\log\left(f(\boldsymbol{x}+\boldsymbol{r},i)\right))-\log\left(f(\boldsymbol{x}+\boldsymbol{r},t)\right),-k\right)\label{eq:-3}
\end{equation}
\\
where the input \textbf{$\boldsymbol{x}$}, correctly classified by
the classifier $f$, is perturbed with \textbf{$\boldsymbol{r}$},
such that the resulting adversarial example is \textbf{$\boldsymbol{x}+\boldsymbol{r}$}.

ZOO uses the symmetric difference quotient to estimate the gradient
and Hessian:

\begin{equation}
\frac{\partial f(\boldsymbol{x})}{\partial x_{i}}\approx\frac{f(x+h*e_{i})-f(x-h*e_{i})}{2h}\label{eq:-4}
\end{equation}

\begin{equation}
\frac{\partial^{2}f(\boldsymbol{x})}{\partial x_{i}^{2}}\approx\frac{f(x+h*e_{i})-2f(x)+f(x-h*e_{i})}{h^{2}}
\end{equation}
\\
where $e_{i}$ denotes the standard basis vector with the i-th component
as 1, and $h$ is a small constant.

Using Equations \ref{eq:-3} and \ref{eq:-4}, the target classifier's
gradient can be numerically derived from the confidence scores of
adjacent input points, and then a gradient-based attack is applied,
in the direction of maximum impact, in order to generate an adversarial
example.

\subsection*{Decision-Based Attacks}

\emph{Decision-based attacks} only use the label predicted by the
target classifier. Thus, these are black-box attacks.

\subsubsection*{Generative Adversarial Network (GAN)}

An adversary can try to generate adversarial examples based on a GAN,
a generative model introduced in \citet{DBLP:conf/nips/GoodfellowPMXWOCB14}.
A GAN is designed to generate fake samples that cannot be distinguished
from the original samples. A GAN is composed of two components: a
discriminator and a generator. The generator is a generative neural
network used to generate samples. The discriminator is a binary classifier
used to determine whether the generated samples are real or fake.
The discriminator and generator are alternately trained so that the
generator can generate valid adversarial records. Assuming we have
the original sample set $x$ with distribution $p_{r}$ and input
noise variables $z$ with distribution $p_{z}$, $G$ is a generative
multilayer perception function with parameter g that generates fake
samples $G(z)$; another model $D$ is a discriminative multilayer
perception function that outputs $D(x)$, which represents the probability
that model $D$ correctly distinguishes fake samples from the original
samples. $D$ and $G$ play the following two player minimax game
with the value function $V(G;D)$:

\begin{equation}
\underset{G}{\min}\underset{D}{\max}V(G,D)=\underset{X\sim p_{r}}{E}\left[log\left(D(X)\right)\right]+\underset{Z\sim p_{z}}{E}\left[log\left(1-D(G(Z)\right)\right]
\end{equation}

In this competing fashion, a GAN is capable of generating raw data
samples that look close to the real data.

\subsubsection*{The Transferability Property}

Many black-box attacks presented in this paper (e.g., \citet{DBLP:conf/raid/RosenbergSRE18,DBLP:journals/corr/abs-1902-08909,Yang18})
rely on the concept of \emph{adversarial example transferability}
presented in \citet{DBLP:journals/corr/SzegedyZSBEGF13}: Adversarial
examples crafted against one model are also likely to be effective
against other models. This transferability property holds even when
the models are trained on different datasets. This means that the
adversary can train a \emph{surrogate model}, which has similar decision
boundaries as the original model and perform a white-box attack on
it. Adversarial examples that successfully fool the surrogate model
would most likely fool the original model as well (\citet{Papernot:2017:PBA:3052973.3053009}). 

The transferability between DNNs and other models, such as decision
tree and SVM models, was examined in \citet{DBLP:journals/corr/PapernotMG16}.
A study of the transferability property using large models and a large-scale
dataset was conducted in \citet{DBLP:journals/corr/LiuCLS16}, which
showed that while transferable non-targeted adversarial examples are
easy to find, targeted adversarial examples rarely transfer with their
target labels. However, for binary classifiers (commonly used in the
cyber security domain), targeted and non-targeted attacks are the
same.

The reasons for the transferability are unknown, but a recent study
\citet{ilyas2019adversarial} suggested that adversarial vulnerability
is not ``necessarily tied to the standard training framework but
is rather a property of the dataset (due to representation learning
of non-robust features)''; this also clarifies why transferability
happens regardless of the classifier architecture. This can also explain
why transferability is applicable to training phase attacks (e.g.,
poisoning attacks) as well (\citet{Munoz-Gonzalez:2017:TPD:3128572.3140451}).

\section{\label{sec:Taxonomy}Taxonomy}

Adversarial learning in cyber security is the modeling of non-stationary
adversarial settings like spam filtering or malware detection, where
a malicious adversary can carefully manipulate (or perturb) the input
data, exploiting specific vulnerabilities of learning algorithms in
order to compromise the (targeted) machine learning system's security. 

A taxonomy for the adversarial domain in general exists (e.g., \citet{DBLP:journals/ml/BarrenoNJT10})
and inspired our taxonomy. However, the cyber security domain has
a few unique challenges, described in the previous section, necessitating
a different taxonomy to categorize the existing attacks, with several
new parts, e.g., the attack's output, attack's targeting, and perturbed
features.

\begin{figure*}[t]
\begin{centering}
\textsf{\includegraphics[scale=0.75]{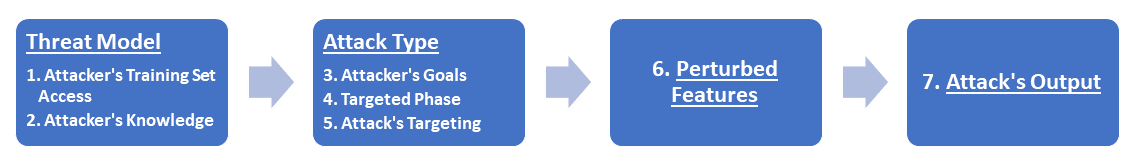}}
\par\end{centering}
\caption{\label{fig:Overview-of-Our-1}Chronological overview of the taxonomy}
\end{figure*}

Our proposed taxonomy is shown in Figure \ref{fig:Overview-of-Our-1}.
The attacks are categorized based on seven distinct attack characteristics,
which are sorted by four chronological phases of the attack:

1) Threat Model - The attacker's knowledge and capabilities, known
prior to the attack. The threat model includes the training set access
and attacker's knowledge.

2) Attack Type - These characteristics are a part of the attack implementation.
The attack type includes the attacker's goals, the targeted phase,
and the attack's targeting.

3) The features modified (or perturbed) by the attack. 

4) The attack's output.

\begin{figure*}[t]
\begin{centering}
\textsf{\includegraphics{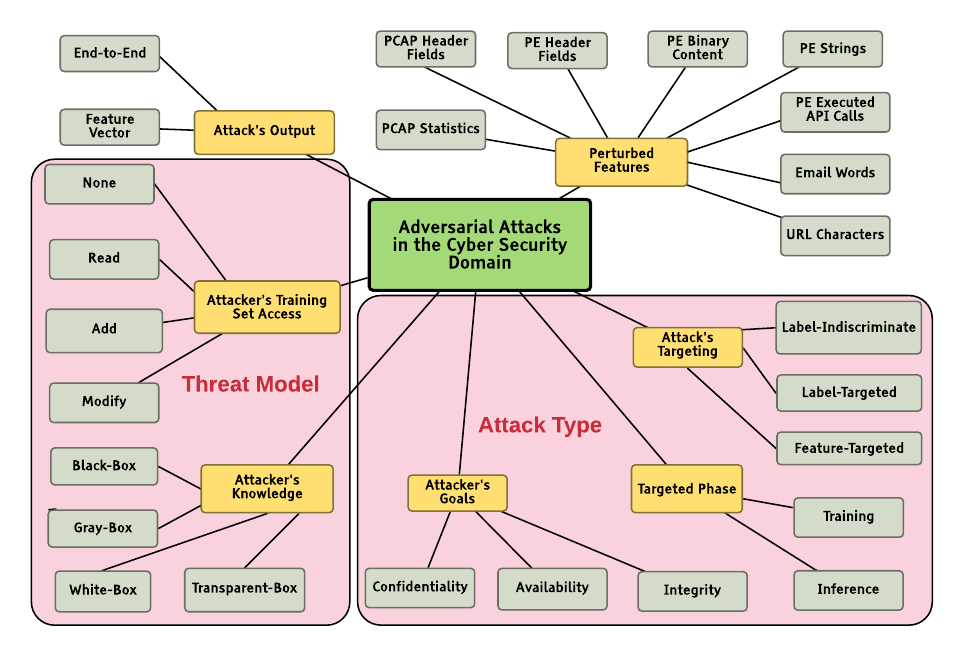}}
\par\end{centering}
\caption{\label{fig:Overview-of-Our}Detailed overview of the taxonomy}
\end{figure*}

A more detailed overview of our proposed taxonomy, including possible
values for the seven characteristics, is shown in Figure \ref{fig:Overview-of-Our}.
The seven attack characteristics (attacker's goals, attacker's knowledge,
attacker's training set access, targeted phase, attack's targeting,
perturbed features, and attack's output) are described in the subsections
that follow.

We include these characteristics in our taxonomy for the following
reasons:

1) These characteristics are specific to the cyber domain (e.g., perturbed
features and attack\textquoteright s output).

2) These characteristics are especially relevant to the threat model,
which plays a much more critical role in the cyber security domain,
where white-box attack are less valuable than in other domains, since
the knowledge of adversaries in the cyber security domain about the
classifier architecture is usually very limited (e.g., attacker\textquoteright s
knowledge, the attacker\textquoteright s training set access, and
the targeted phase).

3) These characteristics highlight missing research in the cyber security
domain, which exists in other domains of adversarial learning. Such
research is specified in Section 7 \ref{sec:Adversarial-Learning-FutureUniqu}
(e.g., attack's targeting).

4) These characteristics exist in many domains but have a different
emphasis (and are therefore more important) in the cyber security
domain (for example, if we analyze the attacker\textquoteright s goal
characteristic, availability attacks are of limited use in other domains,
but they are very relevant in the cyber security domain).

\subsection{\label{subsec:Attacker's-Goals}Attacker's Goals}

This characteristic of the attack is sometimes considered part of
the attack type. An attacker aims to achieve one or more of the following
goals (a.k.a. the CIA triad): (1) \emph{Confidentiality} - Acquire
private information by querying the machine learning system, e.g.,
stealing the classifier's model (\citet{DBLP:conf/uss/TramerZJRR16}),
(2) \emph{Integrity} - Cause the machine learning system to perform
incorrectly for some or all input; for example, to cause a machine
learning-based malware classifier to misclassify a malware sample
as benign (\citet{DBLP:conf/sp/SrndicL14}), and (3)\emph{ Availability}
- Cause a machine learning system to become unavailable or block regular
use of the system; for instance, to generate malicious sessions which
have many of the features of regular traffic, causing the system to
classify legitimate traffic sessions as malicious and block legitimate
traffic (\citet{Chung06}).

\subsection{\label{subsec:Attacker's-Knowledge}Attacker's Knowledge}

This attack characteristic is sometimes considered part of the threat
model. Attacks vary based on the amount of knowledge the adversary
has about the classifier he/she is trying to subvert: (1) \emph{Black-Box
attack} - Requires no knowledge about the model beyond the ability
to query it as a black-box (a.k.a. the oracle model), i.e., inserting
an input and obtaining the output classification, (2) \emph{Gray-Box
attack} - Requires some (limited) degree of knowledge about the targeted
classifier. While usually this consists of the features monitored
by the classifier, sometimes it is other incomplete pieces of information
like the output of the hidden layers of the classifier or the confidence
score (and not just the class label) provided by the classifier, (3)
\emph{White-Box attack} - The adversary has knowledge about the model
architecture and even the hyperparameters used to train the model,
and (4) \emph{Transparent-Box attack} - In this case, the adversary
has complete knowledge about the system, including both white-box
knowledge and knowledge about the defense methods used by defender
(see Section \ref{subsec:Adversarial-Defense-Methods}). Such knowledge
can assist the attacker in choosing an adaptive attack that would
be capable of bypassing the specific defense mechanism (e.g., \citet{tramer2020adaptive}).

While white-box attacks tend to be more efficient than black-box attacks
(sometimes by an order of magnitude \citet{Rosenberg2016}), the knowledge
required is rarely available in real-world use cases. However, white-box
knowledge can be gained either through internal knowledge or by using
a staged attack to reverse engineer the model beforehand \citet{DBLP:conf/uss/TramerZJRR16}.
Each type of attack (black-box, gray-box, etc.) has a query-efficient
variant in which the adversary has only a limited number of queries
(in each query the adversary inserts input into the classifier and
obtains its classification label), and not an unlimited amount of
queries, as in the variants mentioned above. A query-efficient variant
is relevant in the case of cloud security services (e.g., \citet{DBLP:conf/acsac/RosenbergSER20}).
In such services, the attacker pays for every query of the target
classifier and therefore aims to minimize the number of queries made
to the cloud service when performing an attack. Another reason for
minimizing the number of queries is that many queries from the same
computer might arouse suspicion of an adversarial attack attempt,
causing the cloud service to stop responding to those queries. Such
cases require query-efficient attacks.

\subsection{\label{subsec:Attacker's-Capabilities}Attacker's Training Set Access}

Another important characteristic of an attack, sometimes considered
part of the threat model, is the access the adversary has to the training
set used by the classifier (as opposed to access to the model itself,
mentioned in the previous subsection). The attacker\textquoteright s
training set access is categorized as follows: (1) None - no access
to the training set, (2)\emph{ Read} data from the training set (entirely
or partially), (3)\emph{ Add} new samples to the training set, and
(4) \emph{Modify} existing samples (modifying either all features
or just specific features, e.g., the label). For instance, poisoning
attacks require \emph{add} or \emph{modify} permissions.

\subsection{\label{subsec:Attack's-Phase}Targeted Phase}

This attack characteristic is sometimes considered part of the attack
type. Adversarial attacks against machine learning systems occur in
two main phases of the machine learning process: (1)\emph{ Training
Phase attack }- This attack aims to introduce vulnerabilities (to
be exploited in the classification phase) by manipulating training
data during the training phase. For instance, a \emph{poisoning attack}
can be performed by inserting crafted malicious samples labeled as
benign to the training set as part of the baseline training phase
of a classifier, (2)\emph{ Inference Phase attack} - This attack aims
to find and subsequently exploit vulnerabilities in the classification
phase. In this phase, the attacker modifies only samples from the
test set. For example, an \emph{evasion attack} involves modifying
the analyzed malicious sample's features in order to evade detection
by the model. Such inputs are called \emph{adversarial examples}. 

Note that attacks on online learning systems (for instance, anomaly
detection systems \citet{DBLP:journals/corr/abs-1903-11688}) combine
both training phase and inference phase attacks: the attack is an
evasion attack, but if it succeeds, the classifier learns that this
traffic is legitimate, making additional such attacks harder to detect
by the system (i.e., there is a poisoning effect). Such attacks would
be termed inference attacks in this paper, since in this case, the
poisoning aspect is usually a by-product, and is not the attacker's
main goal. Moreover, even if the poisoning aspect is important to
the attacker, it would usually be successful only if the evasion attack
works, so evasion is the primary goal of the attacker in any case.

\subsection{\label{subsec:Attack's-Targeting}Attack's Targeting}

This characteristic is sometimes considered part of the attack type.
Each attack has a different targeting, defining the trigger conditions
or the desired effect on the classifier: (1)\emph{ Label-Indiscriminate
attack} - Always minimizes the probability of correctly classifying
a perturbed sample (the adversarial example), (2) \emph{Label-Targeted
attack} - Always maximizes the probability of a specific class to
be predicted for the adversarial example (different from the predicted
class for the unperturbed sample), and (3)\emph{ Feature-Targeted
attack} - The malicious behavior of these attacks are only activated
by inputs stamped with an attack trigger, which might be the existence
of a specific input feature or group of feature values in the adversarial
example.

Attacks can be both feature and label targeted. Note that in the cyber
security domain, many classifiers are binary (i.e., they have two
output classes: malicious and benign, spam and ham, anomalous or not,
etc.). For binary classifiers, label-indiscriminate and label-targeted
attacks are the same, because in these cases, minimizing the probability
of the current class (label-indiscriminate attack) is equivalent to
maximizing the probability of the only other possible class. 

\subsection{Perturbed Features}

As mentioned in Section \ref{subsec:Executables-are-More}, in the
cyber security domain, classifiers and other machine learning systems
often use more than one feature type. Thus, attackers who want to
subvert those systems should consider modifying more than a single
feature type. We can therefore characterize the different adversarial
attacks in the cyber security domain by the features being modified/perturbed
or added. Note that the same feature type might be modified differently
depending on the sample's format. For instance, modifying a printable
string inside a PE file might be more challenging than modifying a
word within the content of an email content, although the feature
type is the same. Thus, this classification is not simply a feature
type but a tuple of feature type and sample format (for instance,
printable strings inside a PE file). The following is a \emph{partial}
list (e.g., the work of Rosenberg et al. \citep{RosenbergIJCNN2020}
contains 2,381 features, so the full list cannot be included) of such
tuples used in the papers reviewed in our research: PCAP (packet capture;
part of a network session) statistical features (e.g., number of SYN
requests in a certain time window), PCAP header (e.g., IP or UDP)
fields, PE header fields, printable strings inside a PE file, binary
bytes inside a PE file, PE executed API calls (during a dynamic analysis
of the PE file), and words inside an email or characters inside a
URL.

\subsection{\label{subsec:End-to-end-Attacks}Attack's Output}

As discussed in Section \ref{subsec:ExecutableMaintaining-(Malicious},
in contrast to image-based attacks, most adversarial attacks in the
cyber domain require the modification of a feature's values. While
in some domains, such as spam detection, modifying a word in an email
is non-destructive, modifying, e.g., a field in a PE header metadata,
might result in an unrunnable PE file. Thus, there are two type of
attacks: (1)\emph{ Feature-Vector attack} - Such attacks obtain a
feature vector as an input and output another perturbed feature vector.
However, such an attack doesn't generate a sample which can be used
by the attacker and is usually only a hypothetical attack that would
not be possible in real life, and (2)\emph{ End-to-End attack} - This
attack generates a functional sample as an output. Thus, this is a
concrete real-life attack. This category is further divided into many
subgroups based on the sample type produced, e.g., a valid and runnable
PE file, a phishing URL, a spam email, etc.

For instance, most traffic anomaly detection attacks reviewed in this
paper are feature vector attacks. They use statistical features aggregating
packet metadata, but the authors do not show how to generate the perturbed
packet. In contrast, the attack used by \citet{DBLP:conf/raid/RosenbergSRE18}
to add API calls to a malicious process uses a custom framework that
generates a new binary that adds those API calls. Thus, this is an
end-to-end attack. In some image-based domains, e.g., face recognition
systems (Section \ref{subsec:Face-Recognition}), end-to-end attacks
can be further categorized as those that generate images (e.g., \citet{Trojannn})
and those that generate physical elements that can be used to generate
multiple relevant images (e.g., \citet{Sharif:2016:ACR:2976749.2978392}).

\section{\label{sec:Cyber-Applications-of}Adversarial Attacks in the Cyber
Security Domain}

Our paper addresses adversarial attacks in the cyber security domain.
An overview of this section is provided in Tables \ref{tab:Comparison-of-Adversarial-1}-\ref{tab:Comparison-of-Adversarial-6}.
Target classifier abbreviations are specified in Appendix \ref{sec:Appendix-A:-Deep}.
The attack type includes the attacker's goals, the targeted phase,
and the attack's targeting. The threat model includes the attacker's
knowledge and training set access. Unless otherwise mentioned, a gray-box
attack requires knowledge of the target classifier's features, the
attack's targeting is label-indiscriminate, and the attacker's training
set access is \emph{none}. Some of the columns are not a part of our
taxonomy (Section \ref{sec:Taxonomy}) but provide additional relevant
information that may be helpful for understanding the attacks, such
as the target classifiers.

Each of the following subsections represents a specific cyber security
domain that uses adversarial learning and discusses the adversarial
learning methods used in this domain. While there are other domains
in cyber security, we focused only on domains in which substantial
adversarial learning research has been performed. Due to space limits,
this review paper covers only the state of the art in the abovementioned
areas and not all adversarial attacks, especially in large and diverse
domains, such as biometric or cyber-physical systems. The strengths
and weaknesses of each adversarial attack are analyzed throughout
this section.

For the reader's convenience, we have summarized the analysis in Tables
\ref{tab:Comparison-of-Adversarial-1}-\ref{tab:Comparison-of-Adversarial-6}
using the following Boolean parameters:

(1) Reasonable attacker knowledge \textendash{} Is the attack a gray-box
or black-box attack, both of which require a reasonable amount of
knowledge (+ value in Tables \ref{tab:Comparison-of-Adversarial-1}-\ref{tab:Comparison-of-Adversarial-6})
or a white-box attack, which requires an unreasonable amount of knowledge
(- value in Tables \ref{tab:Comparison-of-Adversarial-1}-\ref{tab:Comparison-of-Adversarial-6})
in the cyber security domain?

(2) End-to-end attack \textendash{} Does the attack have an end-to-end
attack output (defined in Section \ref{subsec:End-to-end-Attacks})?
Such attacks are considered more feasible attacks (+ value in Tables
\ref{tab:Comparison-of-Adversarial-1}-\ref{tab:Comparison-of-Adversarial-6}),
while feature vector attacks (defined in Section \ref{subsec:End-to-end-Attacks})
are considered less feasible (- value in Tables \ref{tab:Comparison-of-Adversarial-1}-\ref{tab:Comparison-of-Adversarial-6}).

(3) Effective attack \textendash{} Is the attack effective (success
rate greater than 90\%) for the attack use case (+ value in Tables
\ref{tab:Comparison-of-Adversarial-1}-\ref{tab:Comparison-of-Adversarial-6}),
or ineffective (success rate lower than 90\%; - value in Tables \ref{tab:Comparison-of-Adversarial-1}-\ref{tab:Comparison-of-Adversarial-6})?

(4) Representative dataset - Is the dataset used representative of
the relevant threats in the wild (+ value in Tables \ref{tab:Comparison-of-Adversarial-1}-\ref{tab:Comparison-of-Adversarial-6}),
or is it just a subset (or old variants) of those threats (- value
in Tables \ref{tab:Comparison-of-Adversarial-1}-\ref{tab:Comparison-of-Adversarial-6})?

(5) Representative features - Are the features used in the classifiers
being attacked similar to those being used in real-life security products
(+ value in Tables \ref{tab:Comparison-of-Adversarial-1}-\ref{tab:Comparison-of-Adversarial-6})
or not (- value in Tables \ref{tab:Comparison-of-Adversarial-1}-\ref{tab:Comparison-of-Adversarial-6})?

The mathematical background for the deep learning classifiers is provided
in Appendix \ref{sec:Appendix-A:-Deep}, and the mathematical background
for the commonly used adversarial learning attacks in the cyber security
domain is provided in Section \ref{sec:Adversarial-Learning-Methods}.

Note that while the classifiers the attacker tries to subvert are
mentioned briefly in order to provide context helpful for understanding
the attack, a complete list of the state-of-the-art prior work is
not provided due to space limits. A more comprehensive list can be
found, e.g., in \citet{Berman19}. Cases in which an adversarial attack
does not exist for a specific application type are omitted. This paper
also does not review adversarial attacks in non-cyber domains, such
as image recognition (with the exception of the face recognition domain
which is addressed in Section \ref{subsec:Face-Recognition}, which
is cyber related). It also does not cover papers related to cyber
security, but not to adversarial learning, such as the use of machine
learning to bypass CAPTCHA.

\subsection{Malware Detection and Classification}

Next generation antivirus (NGAV) products, such as \href{https://www.cylance.com/en_us/products/our-products/protect.html}{Cylance},
\href{https://www.crowdstrike.com/wp-content/brochures/Falcon-Prevent-FINAL.pdf}{CrowdStrike},
\href{https://www.sentinelone.com/insights/endpoint-protection-platform-datasheet/}{SentinelOne},
and \href{https://www.microsoft.com/security/blog/2018/02/14/how-artificial-intelligence-stopped-an-emotet-outbreak/}{Microsoft ATP}
use machine and deep learning models instead of signatures and heuristics,
allowing them to detect unseen and unsigned malware but also leaving
them open to attacks against such models.

Malware classifiers can either use static features gathered without
running the code (e.g., n-gram byte sequence, strings, or structural
features of the inspected code) or dynamic features (e.g., CPU usage)
collected during the inspected code execution.

While using static analysis provides a performance advantage, it has
a major disadvantage: since the code is not executed, the analyzed
code might not reveal its ``true features.'' For example, when looking
for specific strings in the file, one might not be able to catch polymorphic
malware, in which those features are either encrypted or packed, and
decrypted only during runtime by a specific bootstrap code. Fileless
attacks (code injection, process hollowing, etc.) are also a problem
for static analysis. Thus, dynamic features, extracted at runtime,
can be used. The most prominent dynamic features that can be collected
during malware execution are the sequences of API calls (\citet{Kolbitsch:2009:EEM:1855768.1855790}),
particularly those made to the OS, which are termed system calls.
Those system calls characterize the software behavior and are harder
to obfuscate during execution time without harming the functionality
of the code. The machine learning techniques (and thus the attacks
of them) can be divided into two groups: traditional (or shallow)
machine learning and deep learning techniques. A summary of the attacks
in the malware detection sub-domain is shown in Tables \ref{tab:Comparison-of-Adversarial-1}
and \ref{tab:Comparison-of-Adversarial-1-2}. 

\begin{table}
\caption{\label{tab:Comparison-of-Adversarial-1}Summary of Adversarial Learning
Approaches in Malware Detection (Part 1)}

\centering{}%
\begin{tabular}{>{\centering}p{0.12\textwidth}>{\centering}p{0.04\textwidth}>{\centering}p{0.08\textwidth}>{\centering}p{0.08\textwidth}>{\centering}p{0.1\textwidth}>{\centering}p{0.05\textwidth}>{\centering}p{0.1\textwidth}>{\centering}p{0.03\textwidth}>{\centering}p{0.03\textwidth}>{\centering}p{0.03\textwidth}>{\centering}p{0.03\textwidth}>{\centering}p{0.03\textwidth}}
Citation & Year & Target Classifier & Attack Type & Attack's Output & Threat Model & Perturbed Features & \rotatebox{90}{Reasonable attacker knowledge?} & \rotatebox{90}{End-to-end attack?} & \rotatebox{90}{Effective attack?} & \rotatebox{90}{Representative dataset?} & \rotatebox{90}{Representative features?}\tabularnewline
\midrule
\citet{DBLP:conf/sp/SrndicL14,li2020feature} & 2020 & RF & Inference integrity & PDF file (end-to-end) & Gray-box & Static structural PDF features & + & + & + & - & +\tabularnewline
\midrule
\citet{DBLP:conf/acns/MingXLW0M15} & 2015 & SCDG & Inference integrity & PE file (end-to-end) & Gray-box & Executed API calls & + & + & + & - & -\tabularnewline
\midrule
\citet{Suciu18} & 2018 & SVM & Training integrity & Feature vector & Gray-box; \emph{add} training set access & Static Android manifest features & - & - & + & - & +\tabularnewline
\midrule
\citet{DBLP:conf/ccs/DangHC17} & 2017 & SVM, RF & Inference integrity & PDF file (end-to-end) & Query-efficient gray-box & Static structural PDF features & + & + & + & - & +\tabularnewline
\midrule
\citet{DBLP:journals/corr/abs-1801-08917} & 2018 & GBDT & Inference integrity & PE file (end-to-end) & Black-box  & Operations (e.g., packing) performed on a PE file & + & + & - & + & +\tabularnewline
\midrule
\citet{Grosse2017} & 2017 & FC DNN & Inference integrity & Feature vector & White-box  & Static Android manifest features & - & - & + & -  & +\tabularnewline
\midrule
\citet{xu2020manis} & 2020 & SCDG & Inference integrity & Feature vector & Gray-box  & Static Android manifest features & + & - & + & -  & -\tabularnewline
\midrule
\citet{DBLP:journals/corr/abs-1802-04528,DBLP:conf/eusipco/KolosnjajiDBMGE18} & 2018 & 1D CNN & Inference integrity & PE file (end-to-end) & White-box  & PE file's raw bytes & - & + & + & +  & +\tabularnewline
\bottomrule
\end{tabular}
\end{table}

\begin{table}
\caption{\label{tab:Comparison-of-Adversarial-1-2}Summary of Adversarial Learning
Approaches in Malware Detection (Part 2)}

\centering{}%
\begin{tabular}{>{\centering}p{0.09\textwidth}>{\centering}p{0.04\textwidth}>{\centering}p{0.15\textwidth}>{\centering}p{0.08\textwidth}>{\centering}p{0.1\textwidth}>{\centering}p{0.05\textwidth}>{\centering}p{0.1\textwidth}>{\centering}p{0.03\textwidth}>{\centering}p{0.03\textwidth}>{\centering}p{0.03\textwidth}>{\centering}p{0.03\textwidth}>{\centering}p{0.03\textwidth}}
Citation & Year & Target Classifier & Attack Type & Attack's Output & Threat Model & Perturbed Features & \rotatebox{90}{Reasonable attacker knowledge?} & \rotatebox{90}{End-to-end attack?} & \rotatebox{90}{Effective attack?} & \rotatebox{90}{Representative dataset?} & \rotatebox{90}{Representative features?}\tabularnewline
\midrule
\citet{DBLP:conf/aaaifs/SuciuCJ18} & 2018 & 1D CNN & Inference integrity & PE file (end-to-end) & Black-box  & PE file's raw bytes & + & + & + & + & +\tabularnewline
\midrule
\citet{RosenbergIJCNN2020,RosenbergBHEU20} & 2020 & GBDT, FC DNN & Inference integrity & PE file (end-to-end) & Black-box  & PE header metadata & + & + & + & + & +\tabularnewline
\midrule
\citet{DBLP:journals/corr/HuT17} & 2017 & RF, LR, DT, SVM, MLP & Inference integrity & Feature vector & Gray-box & API calls' unigrams & + & - & + & - & -\tabularnewline
\midrule
\citet{DBLP:conf/ndss/XuQE16} & 2016 & SVM, RF, CNN & Inference integrity & PDF file (end-to-end) & Gray-box & Static structural PDF features & + & + & + & - & +\tabularnewline
\midrule
\citet{DBLP:journals/sensors/LiuDZZWG19} & 2019 & FC DNN, LR, DT, RF & Inference integrity & Feature vector & Gray-box & Static Android manifest features & + & - & + & - & +\tabularnewline
\midrule
\citet{abusnaina2019adversarial} & 2019 & CNN & Inference integrity & Feature vector & White-box & CFG features & - & - & + & - & -\tabularnewline
\midrule
\citet{DBLP:journals/corr/HuT17a} & 2017 & LSTM & Inference integrity & Feature vector & Gray-box & Executed API calls & + & - & + & - & +\tabularnewline
\midrule
\citet{DBLP:conf/raid/RosenbergSRE18} & 2018 & LSTM, GRU, FC DNN, 1D CNN, RF, SVM, LR, GBDT & Inference integrity & PE file (end-to-end) & Gray-box & Executed API calls, printable strings & + & + & + & + & +\tabularnewline
\midrule
\citet{DBLP:conf/acsac/RosenbergSER20} & 2018 & LSTM, GRU, FC DNN, 1D CNN, RF, SVM, LR, GBDT & Inference integrity & PE file (end-to-end) & Query-efficient gray-box & Executed API calls, printable strings & + & + & + & + & +\tabularnewline
\bottomrule
\end{tabular}
\end{table}

\subsubsection{Attacking Traditional (Shallow) Machine Learning Malware Classifiers}

\citet{DBLP:conf/sp/SrndicL14} implemented an inference integrity
gray-box evasion attack against PDFRATE, a random forest classifier
for static analysis of malicious PDF files, using PDF structural features,
e.g., the number of embedded images or binary streams within the PDF.
The attack used either a mimicry attack in which features were added
to the malicious PDF to make it ``feature-wise similar'' to a benign
sample, or created an SVM representation of the classifier and subverted
it using a method that follows the gradient of the weighted sum of
the classifier\textquoteright s decision function and the estimated
density function of benign examples. This ensures that the final result
lies close to the region populated by real benign examples. The density
function must be estimated beforehand, using the standard technique
of kernel density estimation, and then the transferability property
is used to attack the original PDFRATE classifier using the same PDF
file. \citet{li2020feature} performed an inference integrity gray-box
attack against the same classifier by using GAN-generated feature
vectors and transforming them back into PDF files. The main advantage
of these attacks is the fact that they are end-to-end attacks that
produce a valid (malicious) PDF file, which evade detection. The main
problem is that very few malware have PDF file type. PE files, which
are more common, were not covered in this work.

\citet{DBLP:conf/acns/MingXLW0M15} used an inference integrity replacement
attack, replacing API calls with different functionality-preserving
API subsequences (so gray-box knowledge about the monitored APIs is
required) to modify the malware code. They utilized a system-call
dependence graph (SCDG) with the graph edit distance and Jaccard index
as clustering parameters of different malware variants and used several
SCDG transformations on their malware source code to move it to a
different cluster. Their transformations can cause similar malware
variants to be classified as a different cluster, but they didn't
show that the attack can cause malware to be classified (or clustered)
as a benign program, which is usually the attacker's main goal. \citet{xu2020manis}
also implemented an inference integrity gray-box attack against a
SCDG-based APK malware classifier, using n-strongest nodes and FGSM
(see Section \ref{sec:Adversarial-Learning-Methods}) methods. The
main advantage of these attacks is the fact that they are end-to-end
attacks that produce a valid (malicious) binary (PE or APK, respectively)
file, which evade detection. The main problem is that the features
used (SCDG) are not used by real-life NGAV products.

\citet{Suciu18} and \citet{CHEN2018326} used a training integrity
poisoning attack against a linear SVM classifier trained on the Drebin
dataset \citet{DBLP:conf/ndss/ArpSHGR14} for Android malware detection.
This attack requires gray-box knowledge of the classifier's features
and training set \emph{add} access. The poisoning was done by adding
static features (permissions, API calls, URL requests) from the target
to existing benign instances. \citet{Munoz-Gonzalez:2017:TPD:3128572.3140451}
used a training integrity poisoning attack against logistic regression,
MLP, and ADALINE classifiers, for spam and ransomware detection, by
using back-gradient optimization. This attack requires gray-box knowledge
of the classifier's features and training set \emph{add} and \emph{read}
access. A substitute model is built and poisoned, and the poisoned
samples are effective against the target classifier as well, due to
the transferability property. The main problem with these poisoning
attacks is that they require a powerful attacker who is able to inject
samples into the training set of the malware classifier. While such
a scenario is possible in some cases (e.g., supply chain attacks),
this is usually not a common case, making such attacks less feasible.

\citet{DBLP:conf/ccs/DangHC17} utilized the rate of feature modifications
from a malicious sample and a benign known sample as the score and
used a hillclimbing approach to minimize this score, bypassing SVM
and random forest PDF malware classifiers based on static features
in a query-efficient manner. Thus, their inference integrity attack
is a query-efficient gray-box attack. This attack was the first attempt
to perform a query-efficient attack in the cyber security domain.
However, the classifiers bypassed were not state-of-the-art deep classifiers.

In \citet{DBLP:journals/corr/abs-1801-08917,Anderson17}, the features
used by the gradient boosted decision tree classifier included PE
header metadata, section metadata, and import/export table metadata.
In \citet{DBLP:journals/corr/abs-1801-08917,Anderson17}, an inference
integrity black-box attack which trains a reinforcement learning agent
was presented. The agent is equipped with a set of operations (such
as packing) that it may perform on the PE file. The reward function
was the evasion rate. Through a series of games played against the
target classifier, the agent learns which sequences of operations
are likely to result in detection evasion for any given malware sample.
The perturbed samples that bypassed the classifier were uploaded to
\href{https://www.virustotal.com/}{VirusTotal} and scanned by 65
anti-malware products. Those samples were detected as malicious by
50\% of anti-malware products that detected the original unperturbed
samples. This means that this attack works against real-world security
products (although the authors did not mention which ones were affected).
However, unlike other attacks, this attack\textquoteright s effectiveness
is less than 25\% (as opposed to 90\% for most other adversarial attacks),
showing that additional research is needed in order for this approach
to be practical in real-life use cases.

\subsubsection{\label{subsec:DNN-Adversarial-Examples}Attacking Deep Neural Network
Malware Classifiers}

\citet{RosenbergIJCNN2020,RosenbergBHEU20} used the EMBER dataset
and PE structural features (see Table \ref{tab:Commonly-Used-Datasets})
to train a substitute FC DNN model and used explainability machine
learning algorithms (e.g., integrated gradients) to detect which of
the 2,381 features have high impact on the malware classification
and can also be modified without harming the executable's functionality
(e.g., file timestamp). These features were modified in a gray-box
inference integrity attack, and the mutated malware bypassed not only
the substitute model, but also the target GBDT classifier, which used
a different subset of samples and features. The main advantage of
these attacks are that they bypassed an actual, targeted, real-life
NGAV. The main limitation of this attack is that it is not fully automatic
- human intervention is required to select the features that are perturbed.

\citet{Grosse2017,Grosse16} presented a white-box inference integrity
attack against an Android static analysis fully connected DNN malware
classifier. The static features used in the DREBIN dataset (see Table
\ref{tab:Commonly-Used-Datasets}) were from the AndroidManifest.xml
file and included permissions, suspicious API calls, activities, etc.
The attack is a discrete FGSM (see Section \ref{sec:Adversarial-Learning-Methods})
variant, which is performed iteratively in the following two steps
until a benign classification is made: (1) compute the gradient of
the white-box model with respect to the binary feature vector $\boldsymbol{x}$,
and (2) find the element in $\boldsymbol{x}$ whose modification from
zero to one (i.e., only feature addition and not removal) would cause
the maximum change in the benign score and add this feature to the
adversarial example. The main advantage of this attack is that it
provides a methodical way of dealing with discrete features, commonly
used in the cyber security domain, and evaluates this attack against
many DNN architectures. The main problem is that the white-box assumption
is unrealistic in many real-life scenarios.

\citet{DBLP:journals/corr/abs-1802-04528} implemented an inference
integrity attack against MalConv, a 1D CNN, using the file's raw byte
content as features (\citet{DBLP:conf/aaai/RaffBSBCN18}). The additional
bytes are selected by the FGSM method (see Section \ref{sec:Adversarial-Learning-Methods})
and are inserted between the file's sections. \citet{DBLP:conf/eusipco/KolosnjajiDBMGE18}
implemented a similar attack and also analyzed the bytes which are
the most impactful features (and are therefore added by the attack),
showing that a large portion of them are part of the PE header metadata.
\citet{DBLP:conf/aaaifs/SuciuCJ18} transformed this white-box gradient-based
attack to a black-box decision-based attack by appending bytes from
the beginning of benign files, especially from their PE headers, which,
as shown in \citet{DBLP:conf/eusipco/KolosnjajiDBMGE18}, are prominent
features. The main insight of these attacks is that even classifiers
that use raw memory bytes as features (leveraging deep architecture's
representation learning) are vulnerable to adversarial examples. The
main disadvantages is that such classifiers are rarely used by real-life
NGAVs.

\citet{DBLP:journals/corr/HuT17} perturbed static API call unigrams
by performing a gray-box inference integrity attack. If $n$ API types
are used, the feature vector dimension is $n$. A generative adversarial
network (GAN; Appendix \ref{sec:Appendix-A:-Deep}) was trained, where
the discriminator simulates the malware classifier while the generator
tries to generate adversarial samples that would be classified as
benign by the discriminator, which uses labels from the black-box
model (a random forest, logistic regression, decision tree, linear
SVM, MLP, or an ensemble of all of these). However, this is a feature
vector attack: the way to generate an executable with the perturbed
API call trace was not presented, making this attack infeasible in
real life. 

\citet{DBLP:conf/ndss/XuQE16} generated adversarial examples that
bypass PDF malware classifiers, by modifying static PDF features.
This was done using an inference integrity genetic algorithm (GA),
where the fitness of the genetic variants is defined in terms of the
target classifier\textquoteright s confidence score. The GA is computationally
expensive and was evaluated against SVM, random forest, and CNN classifiers
using static PDF structural features. This gray-box attack requires
knowledge of both the classifier's features and the target classifier's
confidence score. \citet{DBLP:journals/sensors/LiuDZZWG19} used the
same approach to bypass an IoT Android malware detector. The bypassed
fully connected DNN, logistic regression, decision tree, and random
forest classifiers were trained using the DREBIN dataset.

\citet{abusnaina2019adversarial} trained an IoT malware detection
CNN classifier using graph-based features (e.g., shortest path, density,
number of edges and nodes, etc.) from the control flow graph (CFG)
of the malware disassembly. They used white-box attacks: C\&W, DeepFool,
FGSM, JSMA (see Section \ref{sec:Adversarial-Learning-Methods}),
the momentum iterative method (MIM), projected gradient descent (PGD),
and virtual adversarial method (VAM). They also added their own attack,
graph embedding, and augmentation, which adds a CFG of a benign sample
to the CFG of a malicious sample via source code concatenation. The
problem with this attack is that CFG takes a long time to generate,
and therefore graph-based features are rarely used by real-life malware
classifiers.

\citet{DBLP:journals/corr/HuT17a} proposed a gray-box inference integrity
attack using an RNN GAN to generate invalid APIs and inserted them
into the original API sequences to bypass an LSTM classifier trained
on the API call trace of the malware. A substitute RNN is trained
to fit the targeted RNN. Gumbel-Softmax, a one-hot continuous distribution
estimator, was used to smooth the API symbols and deliver gradient
information between the generative RNN and the substitute RNN. Null
APIs were added, but while they were omitted to make the adversarial
sequence generated shorter, they remained in the loss function's gradient
calculation. This decreases the attack's effectiveness, since the
substitute model is used to classify the Gumbel-Softmax output, including
the null APIs' estimated gradients, so it does not simulate the malware
classifier exactly. The gray-box attack output is a feature vector
of the API call sequence that might harm the malware's functionality
(e.g., by inserting the \emph{ExitProcess()} API call in the middle
of the malware code), making this attack infeasible in real-life scenarios.

\citet{DBLP:conf/raid/RosenbergSRE18} presented a gray-box inference
integrity attack that adds API calls to an API call trace used as
input to an RNN malware classifier, in order to bypass a classifier
trained on the API call trace of the malware. A GRU substitute model
was created and attacked, and the transferability property was used
to attack the original classifier. The authors extended their attack
to hybrid classifiers combining static and dynamic features, attacking
each feature type in turn. The target models were LSTM variants, GRUs,
conventional RNNs, bidirectional and deep variants, and non-RNN classifiers
(including both feedforward networks, like fully connected DNNs and
1D CNNs, and traditional machine learning classifiers, such as SVM,
random forest, logistic regression, and gradient boosted decision
tree). The authors presented an end-to-end framework that creates
a new malware executable without access to the malware source code.
The variety of classifiers and the end-to-end framework fits real-life
scenarios, but the focus only on strings' static features is limiting.

A subsequent work (\citet{DBLP:conf/acsac/RosenbergSER20}) presented
two query-efficient gray-box inference integrity attacks against the
same classifiers, based on \emph{benign perturbations} generated using
a GAN that was trained on benign samples. One of the gray-box attack
variants requires the adversary to know which API calls are being
monitored, and the other one also requires the confidence score of
the target classifier in order to operate an evolutionary algorithm
to optimize the perturbation search and reduce the number of queries
used. This attack is generic for every camouflaged malware and does
not require a per malware pre-deployment phase to generate the adversarial
sequence (either using a GAN, as in \citet{DBLP:journals/corr/HuT17a},
or a substitute model, as in \citet{DBLP:conf/raid/RosenbergSRE18}).
Moreover, the generation is done at runtime, making it more generic
and easier to deploy.

\subsection{URL Detection}

Webpages are addressed by a uniform resource locator (URL). A URL
begins with the protocol used to access the page. The fully qualified
domain name (FQDN) identifies the server hosting the webpage. It consists
of a registered domain name (RDN) and prefix referred to as subdomains.
A phisher has full control of the subdomain and can set them to any
value. The RDN is constrained, since it has to be registered with
a domain name registrar. The URL may also have a path and query components
which also can be changed by the phisher at will.

Consider this URL example: https://www.amazon.co.uk/ap/signin?encoding=UTF8.
We can identify the following components: protocol = https; FQDN =
www.amazon.co.uk; RDN = amazon.co.uk; path and query = /ap/signin?encoding=UTF8.
A summary of the attacks in the URL detection sub domain is shown
in Table \ref{tab:Comparison-of-Adversarial-2}. 

\begin{table}
\caption{\label{tab:Comparison-of-Adversarial-2}Summary of Adversarial Learning
Approaches in URL Detection}

\centering{}%
\begin{tabular}{>{\centering}p{0.09\textwidth}>{\centering}p{0.04\textwidth}>{\centering}p{0.15\textwidth}>{\centering}p{0.08\textwidth}>{\centering}p{0.1\textwidth}>{\centering}p{0.05\textwidth}>{\centering}p{0.1\textwidth}>{\centering}p{0.03\textwidth}>{\centering}p{0.03\textwidth}>{\centering}p{0.03\textwidth}>{\centering}p{0.03\textwidth}>{\centering}p{0.03\textwidth}}
Citation & Year & Target Classifier & Attack Type & Attack's Output & Threat Model & Perturbed Features & \rotatebox{90}{Reasonable attacker knowledge?} & \rotatebox{90}{End-to-end attack?} & \rotatebox{90}{Effective attack?} & \rotatebox{90}{Representative dataset?} & \rotatebox{90}{Representative features?}\tabularnewline
\midrule
\citet{Bahnsen2018DeepPhishS} & 2018 & LSTM & Inference integrity & URL (end-to-end) & Gray-box & URL characters & + & + & + & + & -\tabularnewline
\midrule
\citet{Shirazi19} & 2019 & State-of-the-art phishing classifiers & Inference integrity & Feature vector & Gray-box & All features used by the classifiers & + & - & + & + & +\tabularnewline
\midrule
\citet{aleroud2020bypassing} & 2020 & RF, NN, DT, LR, SVM & Inference integrity & URL (end-to-end) & Black-box & URL characters & + & + & + & - & -\tabularnewline
\midrule
\citet{DBLP:conf/ccs/AndersonWF16} & 2016 & RF & Inference integrity & URL (end-to-end) & Black-box & URL characters & + & + & + & + & -\tabularnewline
\midrule
\citet{DBLP:journals/corr/abs-1902-08909} & 2019 & CNN, LSTM, BLSTM & Inference integrity & URL (end-to-end) & Black-box & URL characters & + & + & + & - & -\tabularnewline
\bottomrule
\end{tabular}
\end{table}

Since URLs can be quite long, URL shortening services have started
to appear. In addition to shortening the URL, these services also
obfuscate them.

\subsubsection{Phishing URL Detection}

Phishing refers to the class of attacks where a victim is lured to
a fake web page masquerading as a target website and is deceived into
disclosing personal data or credentials. Phishing URLs seem like legitimate
URLs and redirect the users to phishing web pages, which mimic the
look and feel of their target websites (e.g., a bank website), in
the hopes that the user will enter his/her personal information (e.g.,
password).

\citet{Bahnsen2018DeepPhishS} performed an inference integrity attack
to evade a character-level LSTM-based phishing URL classifier (\citet{Bahnsen17}),
by concatenating the effective URLs from historical attacks (thus,
this is a gray-box attack). Then, from this full text, sentences with
a fixed length were created. An LSTM model used those sentences as
a training set in order to generate the next character. Once the model
generated a full prose text, it was divided by http structure delimiters
to produce a list of pseudo URLs. Each pseudo URL was assigned a compromised
domain, such that the synthetic URLs take the form: http://+compromised\_domain+pseudo\_URL.
While this attack is indeed effective, the concatenation of benign
URLs can be signed, making this attack less evasive for real-life
classifiers then the generative attacks (e.g., \citet{aleroud2020bypassing})
mentioned below.

\citet{Shirazi19} generated adversarial examples using all possible
combinations of the values of the features (e.g., website reputation)
used by state-of-the-art phishing classifiers, such as \citet{Verma:2015:CPU:2699026.2699115},
making this a more realistic attack. However, the attack requires
knowledge about the features being used by the classifier, making
it a gray-box inference integrity attack. Such knowledge is not always
accessible to the attacker, making this attack less feasible in real-life
scenarios.

Phishing URLs were generated by a text GAN in \citet{Trevisan:2019:RUC:3308897.3308959,Anand18},
in order to augment the phishing URL classifier's training set and
improve its accuracy. AlEroud and Karabatis \citet{aleroud2020bypassing}
used the phishing URLs generated as adversarial examples in an inference
integrity attack, in order to bypass the target classifier. It remains
unclear whether the attack mentioned in \citet{aleroud2020bypassing}
is robust enough to bypass GAN-based defenses, such as the defense
methods presented in \citet{Trevisan:2019:RUC:3308897.3308959,Anand18}.

\subsubsection{Domain Generation Algorithm (DGA) URL Detection}

DGAs are commonly used malware tools that generate large numbers of
domain names that can be used for difficult to track communications
with command and control servers operated by the attacker. The large
number of varying domain names makes it difficult to block malicious
domains using standard techniques such as blacklisting or sinkholing.
DGAs are used in a variety of cyber attacks, including ransomware,
spam campaigns, theft of personal data, and implementation of distributed
denial-of-service (DDoS) attacks. DGAs allow malware to generate any
number of domain names daily, based on a seed that is shared by the
malware and the threat actor, allowing both to synchronize the generation
of domain names.

\citet{DBLP:journals/corr/abs-1902-08909} used a black-box inference
integrity attack, training a substitute model to simulate the DGA
classifier on a list of publicly available DGA URLs. Then that attacker
iterates over every character in the DGA URL. In each iteration, the
results of the feedforward pass of the substitute model are used to
compute the loss with regard to the benign class. The attacker performs
a single backpropagation step on the loss in order to acquire the
Jacobian-based saliency map, which is a matrix that assigns every
feature in the input URL with a gradient value (JSMA; see Section
\ref{sec:Adversarial-Learning-Methods}). Features (characters) with
higher gradient values in the JSMA would have a more significant (salient)
effect on the misclassification of the input, and thus each character
would be modified in turn, making the substitute model's benign score
higher. Finally, URLs that evade detection by the substitute model
would also evade detection by the target DGA classifier due to the
transferability property (see Section \ref{sec:Adversarial-Learning-Methods}).
Despite the fact that this attack confirms the feasibility of transferability
in the DGA URL detection sub-domain, the use of (only) character-level
features does not accurately represent real-life classifiers.

\citet{DBLP:conf/ccs/AndersonWF16} performed an inference integrity
black-box attack that used a GAN to produce domain names that current
DGA classifiers would have difficulty identifying. The generator was
then used to create synthetic data on which new models were trained.
This was done by building a neural language architecture, a method
of encoding language in a numerical format, using LSTM layers to act
as an autoencoder. The autoencoder is then repurposed such that the
encoder (which takes in domain names and outputs an embedding that
converts a language into a numerical format) acts as the discriminator,
and the decoder (which takes in the embedding and outputs the domain
name) acts as the generator. Anderson et al. attacked a random forest
classifier trained on features defined in \citet{10.1007/978-3-319-08509-8_11,Antonakakis12,Yadav12,Yadav:2010:DAG:1879141.1879148}.
The features of the random forest DGA classifier are unknown to the
attacker. They include: the length of domain name; entropy of character
distribution in domain name; vowel to consonant ratio; Alexa top 1M
n-gram frequency distribution co-occurrence count, where n = 3, 4,
or 5; n-gram normality score; and meaningful character ratio. The
fact that this attack bypasses a classifier which uses many features,
as specified above, makes it more suitable for real-life scenarios.

\subsection{\label{subsec:Network-Intrusion-Detection}Network Intrusion Detection}

A security system commonly used to secure networks is a network intrusion
detection system (NIDS). An NIDS is a device or software that monitors
all traffic passing a strategic point for malicious activities. When
such an activity is detected, an alert is generated. Typically an
NIDS is deployed at a single point, for example, at the Internet gateway.
A summary of the attacks in the network intrusion detection sub-domain
is provided in Table \ref{tab:Comparison-of-Adversarial-3}. 

\begin{table}
\caption{\label{tab:Comparison-of-Adversarial-3}Summary of Adversarial Learning
Approaches in Network Intrusion Detection}

\centering{}%
\begin{tabular}{>{\centering}p{0.09\textwidth}>{\centering}p{0.04\textwidth}>{\centering}p{0.15\textwidth}>{\centering}p{0.1\textwidth}>{\centering}p{0.1\textwidth}>{\centering}p{0.05\textwidth}>{\centering}p{0.1\textwidth}>{\centering}p{0.03\textwidth}>{\centering}p{0.03\textwidth}>{\centering}p{0.03\textwidth}>{\centering}p{0.03\textwidth}>{\centering}p{0.03\textwidth}}
Citation & Year & Target Classifier & Attack Type & Attack's Output & Threat Model & Perturbed Features & \rotatebox{90}{Reasonable attacker knowledge?} & \rotatebox{90}{End-to-end attack?} & \rotatebox{90}{Effective attack?} & \rotatebox{90}{Representative dataset?} & \rotatebox{90}{Representative features?}\tabularnewline
\midrule
\citet{DBLP:journals/corr/abs-1903-11688} & 2019 & Autoencoder ensemble & Inference integrity & Feature vector & White-box & Protocol statistical features & - & - & + & + & +\tabularnewline
\midrule
\citet{DBLP:journals/corr/abs-1809-02077} & 2018 & SVM, NB, MLP, LR, DT, RF, KNN & Inference integrity & Feature vector & Gray-box & Statistical and protocol header features & + & - & + & - & +\tabularnewline
\midrule
\citet{Yang18} & 2018 & DNN & Inference integrity & Feature vector & Gray-box & Same as Lin et al. & + & - & + & - & +\tabularnewline
\midrule
\citet{Rigaki2017AdversarialDL} & 2017 & DT, RF, SVM & Inference integrity & Feature vector & Gray-box & Same as Lin et al. & + & - & + & - & +\tabularnewline
\midrule
\citet{DBLP:conf/inista/WarzynskiK18} & 2018 & MLP & Inference integrity & Feature vector & White-box & Same as Lin et al. & - & - & + & - & +\tabularnewline
\midrule
\citet{Wang18}  & 2018  & MLP & Inference integrity & Feature vector & White-box & Same as Lin et al. & - & - & + & - & +\tabularnewline
\midrule
\citet{Kuppa:2019:BBA:3339252.3339266} & 2019 & DAGMM, AE, AnoGAN, ALAD, DSVDD, OC-SVM, IF & Inference integrity & PCAP file (end-to-end) & Query-efficient gray-box & Similar to Lin et al., but modifies only non-impactful features like
send time & + & + & + & + & +\tabularnewline
\midrule
\citet{DBLP:journals/corr/abs-1905-05137} & 2019 & FC DNN, SNN & Inference integrity & Feature vector & Gray-box & Statistical and protocol header features & + & - & + & + & +\tabularnewline
\midrule
\citet{10.1007/978-981-13-1059-1_17} & 2019 & MLP, CNN, LSTM & Inference integrity, training availability & Feature vector & White-box & Features from SDN messages & - & - & + & + & +\tabularnewline
\bottomrule
\end{tabular}
\end{table}

\citet{DBLP:journals/corr/abs-1903-11688} conducted a white-box inference
integrity attack against Kitsune (\citet{DBLP:conf/ndss/MirskyDES18}),
an ensemble of autoencoders used for online network intrusion detection.
Kitsune uses packet statistics which are fed into a feature mapper
that divides the features between the autoencoders, to ensure fast
online training and prediction. The RMSE output of each autoencoder
is fed into another autoencoder that provides the final RMSE score
used for anomaly detection. This architecture can be executed on small,
weak routers.

Clements et al. used four adversarial methods: FGSM, JSMA, C\&W, and
ENM (see Section \ref{sec:Adversarial-Learning-Methods}). The attacker
uses the $L_{P}$ distance on the feature space between the original
input and the perturbed input as the distance metric. Minimizing the
$L_{0}$ norm correlates to altering a small number of extracted features.
This method has two main limitations: (1) The threat model assumes
that the attacker knows the target classifier's features, architecture,
and hyperparameters. This makes this attack a white-box attack, rather
than black-box attack. This is a less realistic assumption in real-life
scenarios. (2) Modification is done at the feature level (i.e., modifying
only the feature vector) and not at the sample level (i.e., modifying
the network stream). This means that there is no guarantee that those
perturbations can be performed without harming the malicious functionality
of the network stream. The fact that some of the features are statistical
makes the switch from vector modification to sample modification even
more challenging.

\citet{DBLP:journals/corr/abs-1809-02077} generated adversarial examples
using a GAN, called IDSGAN, in which the GAN's discriminator obtains
the labels from the black-box target classifier. The adversarial examples
are evaluated against several target classifiers: SVM, naive Bayes,
MLP, logistic regression, decision tree, random forest, and k-nearest
neighbors classifiers. This attack assumes knowledge about the target
classifier's features, making it a gray-box inference integrity attack.
The features include individual TCP connection features (e.g., the
protocol type), domain knowledge-based features (e.g., a root shell
was obtained), and statistical features of the network sessions (like
the percentage of connections that have SYN errors within a time window).
All features are extracted from the network stream (the NSL-KDD dataset
was used; see Table \ref{tab:Commonly-Used-Datasets}). This attack
generates a statistical feature vector, but the authors do not explain
how to produce a real malicious network stream that has those properties. 

\citet{Yang18} trained a DNN model to classify malicious behavior
in a network using the same features as \citet{DBLP:journals/corr/abs-1809-02077},
achieving performance comparable to state-of-the-art NIDS classifiers,
and then showed how to add small perturbations to the original input
to lead the model to misclassify malicious network packets as benign
while maintaining the maliciousness of these packets, this attack
assumes that an adversary without internal information on the DNN
model is trying to launch black-box attack. Three different black-box
attacks were attempted by the adversary: an attack based on zeroth
order optimization (ZOO; see Section \ref{sec:Adversarial-Learning-Methods}),
an attack based on a GAN (similar to the one proposed by \citet{DBLP:journals/corr/abs-1809-02077}),
and an attack on which a substitute model is trained which is followed
by a C\&W attack (see Section \ref{sec:Adversarial-Learning-Methods})
performed against the substitute model. Applying the generated adversarial
example against the target classifier succeeds due to the transferability
property (see Section \ref{sec:Adversarial-Learning-Methods}). This
paper has the same limitations as \citet{DBLP:journals/corr/abs-1809-02077}:
this gray-box inference integrity attack assumes knowledge about the
target classifier's features and also generates only the feature vectors
and not the samples themselves.

In their gray-box inference integrity attack, Rigaki and Elragal \citet{Rigaki2017AdversarialDL}
used the same NSL-KDD dataset (see Table \ref{tab:Commonly-Used-Datasets}).
Both FGSM and JSMA (see Section \ref{sec:Adversarial-Learning-Methods})
attacks were used to generate adversarial examples against an MLP
substitute classifier, and the results were evaluated against decision
tree, random forest, and linear SVM classifiers. This paper has the
same limitations as \citet{DBLP:journals/corr/abs-1809-02077}: this
attack assumes knowledge about the target classifier's features and
also generates only the feature vectors and not the samples themselves.

\citet{DBLP:conf/inista/WarzynskiK18} performed a white-box inference
integrity feature vector attack against an MLP classifier trained
on the NSL-KDD dataset (see Table \ref{tab:Commonly-Used-Datasets}).
They used a white-box FGSM attack (see Section \ref{sec:Adversarial-Learning-Methods}).
\citet{Wang18} added three more white-box attacks: JSMA, DeepFool,
and C\&W (see Section \ref{sec:Adversarial-Learning-Methods}). The
$L_{p}$ distance and the perturbations are in the feature space in
both cases. This attack is not an end-to-end attack, so again it cannot
be used to generate malicious network streams that bypass real-life
classifiers.

\citet{Kuppa:2019:BBA:3339252.3339266} proposed a query-efficient
gray-box inference integrity attack against deep unsupervised anomaly
detectors, which leverages a manifold approximation algorithm for
query reduction and generates adversarial examples using spherical
local subspaces while limiting the input distortion and KL divergence.
Seven state-of-the-art anomaly detectors with different underlying
architectures were evaluated: a deep autoencoding Gaussian mixture
model, an autoencoder, anoGAN, adversarially learned anomaly detection,
deep support vector data description, one-class support vector machines,
and isolation forests (see Section \ref{sec:Adversarial-Learning-Methods}),
which is a more diverse set of classifiers than other attacks mentioned
in this section, making this attack more applicable, regardless of
the classifier deployed. All classifiers were trained on the CSE-CIC-IDS2018
dataset and features (see Table \ref{tab:Commonly-Used-Datasets}).
This dataset is more recent than the NSL-KDD dataset used in much
of the research mentioned in this section and better represent today's
threats. Unlike other papers discussed in this section, the authors
generated a full PCAP file (and not just feature vectors). They also
only modified features that could be modified without harming the
network stream (e.g., time-based features), so they actually created
adversarial samples and not just feature vectors. However, they did
not run the modified stream in order to verify that the malicious
functionality remains intact. 

\citet{DBLP:journals/corr/abs-1905-05137} attacked a fully connected
DNN and a self-normalizing neural network classifier (an SNN is a
DNN with a SeLU activation layer; \citet{DBLP:conf/nips/KlambauerUMH17})
trained on the BoT-IoT dataset and features (see Table \ref{tab:Commonly-Used-Datasets}),
using FGSM (see Section \ref{sec:Adversarial-Learning-Methods}),
the basic iteration method, and the PGD at the feature level. They
showed that both classifiers were vulnerable, although SNN was more
robust to adversarial examples. This attack is unique in terms of
the dataset and architectures used and demonstrates the susceptibility
of IoT SNN-based classifiers to adversarial attacks.

\citet{10.1007/978-981-13-1059-1_17} attacked port scanning detectors
in a software-defined network (SDN). The detectors were MLP, CNN,
and LSTM classifiers trained on features extracted from Packet-In
messages (used by port scanning tools like Nmap in the SDN) and switch
monitoring statistic messages (STATS). The white-box inference integrity
attacks used were FGSM and JSMA (see Section \ref{sec:Adversarial-Learning-Methods}).
The JSMA attack was also (successfully) conducted on \textbf{regular}
traffic packets (JSMA reverse) to create false negatives, creating
noise and confusion in the network (a white-box training availability
attack). While this attack requires permissions not available to a
regular attacker (knowledge of the classifier's architecture, etc.),
it shows the susceptibility of port scanning classifiers to adversarial
attacks.

\subsection{Spam Filtering}

The purpose of a spam filter is to determine whether an incoming message
is legitimate (i.e., ham) or unsolicited (i.e., spam). Spam detectors
were among the first applications to use machine learning in the cyber
security domain and therefore were the first to be attacked. A summary
of the attacks in the spam filtering sub domain is shown in Table
\ref{tab:Comparison-of-Adversarial-4}. 

\begin{table}
\caption{\label{tab:Comparison-of-Adversarial-4}Summary of Adversarial Learning
Approaches in Spam Filtering}

\centering{}%
\begin{tabular}{>{\centering}p{0.09\textwidth}>{\centering}p{0.04\textwidth}>{\centering}p{0.15\textwidth}>{\centering}p{0.08\textwidth}>{\centering}p{0.1\textwidth}>{\centering}p{0.08\textwidth}>{\centering}p{0.1\textwidth}>{\centering}p{0.03\textwidth}>{\centering}p{0.03\textwidth}>{\centering}p{0.03\textwidth}>{\centering}p{0.03\textwidth}>{\centering}p{0.03\textwidth}}
Citation & Year & Target Classifier & Attack Type & Attack's Output & Threat Model & Perturbed Features & \rotatebox{90}{Reasonable attacker knowledge?} & \rotatebox{90}{End-to-end attack?} & \rotatebox{90}{Effective attack?} & \rotatebox{90}{Representative dataset?} & \rotatebox{90}{Representative features?}\tabularnewline
\midrule
\citet{SETHI2018129}  & 2018  & SVM, kNN, DT, RF & Inference integrity and confidentiality & Feature vector & Gray-box & Email words or same as Lin et al. & + & - & + & + & +\tabularnewline
\midrule
\citet{DBLP:conf/ccs/HuangJNRT11,DBLP:conf/nsdi/NelsonBCJRSSTX08} & 2011 & Bayesian spam filter & Training availability & Email (end-to-end) & Gray-box & Email words & + & + & + & + & +\tabularnewline
\midrule
\citet{Biggio14} & 2014 & SVM, LR & Inference integrity & Email (end-to-end) & White-box & Email words & - & + & + & + & +\tabularnewline
\midrule
\citet{Bruckner:2012:SPG:2503308.2503326} & 2012 & NB, SVM & Inference integrity & Email (end-to-end) & Gray-box & Email words & + & + & + & + & +\tabularnewline
\midrule
\citet{kuleshov2018adversarial} & 2018 & NB, LSTM, 1D CNN & Inference integrity & Email (end-to-end) & Gray-box & Email words & + & + & + & + & +\tabularnewline
\midrule
\citet{DBLP:journals/corr/abs-1812-00151} & 2018 & LSTM, 1D CNN & Inference integrity & Email (end-to-end) & Gray-box & Email words & + & + & + & + & +\tabularnewline
\midrule
\citet{Munoz-Gonzalez:2017:TPD:3128572.3140451} & 2017 & LR, MLP & Training integrity & Feature vector & Gray-box; \emph{add} and \emph{read} training set access & Email words & - & - & + & + & +\tabularnewline
\bottomrule
\end{tabular}
\end{table}

\citet{DBLP:conf/ccs/HuangJNRT11} attacked SpamBayes \citet{Robinson03},
which is a content-based statistical spam filter that classifies email
using token counts. SpamBayes computes a spam score for each token
in the training corpus based on its occurrence in spam and non-spam
emails. The filter computes a message\textquoteright s overall spam
score based on the assumption that the token scores are independent,
and then it applies Fisher\textquoteright s method for combining significance
tests to determine whether the email\textquoteright s tokens are sufficiently
indicative of one class or the other. The message score is compared
against two thresholds to select the label: spam, ham (i.e., non-spam),
or unsure.

\citet{DBLP:conf/ccs/HuangJNRT11} designed two types of training
availability attacks. The first is an indiscriminate dictionary attack,
in which the attacker sends attack messages that contain a very large
set of tokens\textemdash the attack\textquoteright s dictionary. After
training on these attack messages, the victim\textquoteright s spam
filter will have a higher spam score for every token in the dictionary.
As a result, future legitimate email is more likely to be marked as
spam, since it will contain many tokens from that lexicon. The second
attack is a targeted attack - the attacker has some knowledge of a
specific legitimate email he/she targets for incorrect filtering.
\citet{DBLP:conf/nsdi/NelsonBCJRSSTX08} modeled this knowledge by
letting the attacker know a certain fraction of tokens from the target
email, which are included in the attack message. Availability attacks
like these are quite rare in the adversarial learning landscape and
open up interesting attack options, such as adversarial denial-of-service
attacks (see Section \ref{subsec:Attacker's-Goals-Gap:}).

\citet{Biggio14} evaluated the robustness of linear SVM and logistic
regression classifiers to a white-box inference integrity attack where
the attacker adds the most impactful good words and found that while
both classifiers have the same accuracy for unperturbed samples, the
logistic regression classifier outperforms the SVM classifier in robustness
to adversarial examples. The white-box approach used in this attack
is less feasible in real-life scenarios.

\citet{Bruckner:2012:SPG:2503308.2503326} modeled the interaction
between the defender and the attacker in the spam filtering domain
as a static game in which both players act simultaneously, i.e., without
prior information about their opponent\textquoteright s move. When
the optimization criterion of both players depends not only on their
own action but also on their opponent\textquoteright s move, the concept
of a player\textquoteright s optimal action is no longer well-defined,
and thus the cost functions of the learner (the defender) and the
data generator (the attacker) are not necessarily antagonistic. The
authors identified the conditions under which this prediction game
has a unique Nash equilibrium and derived algorithms that find the
equilibrial prediction model. From this equation, they derived new
equations for the Nash logistic regression and Nash SVM using custom
loss functions. The authors showed that both the attacker and the
defender are better off respectively attacking and using the Nash
classifiers. While this attack took a different (more game theory
focused) approach then other more ``conventional'' attacks, its
effectiveness is no different.

\citet{SETHI2018129} trained several classifiers (linear SVM, k-nearest
neighbors, SVM with RBF kernel, decision tree, and random forest)
on several datasets, including SPAMBASE for spam detection and NSL-KDD
(see Table \ref{tab:Commonly-Used-Datasets}) for network intrusion
detection. They presented a gray-box inference integrity and confidentiality
attack and a query-efficient gray-box anchor point inference integrity
attack which is effective against all models. The uniqueness of this
study lies in its focus on query-efficiency (thus making this attack
more feasible) and on the confidentiality attack (trying to reverse
engineer the attacked model, i.e., a model inversion attack), which
are quite rare in the spam filtering domain. 

Generalized attack methods, which are effective against several NLP
classifiers, are a recent trend. \citet{kuleshov2018adversarial}
implemented such a generalized black-box inference integrity attack
to evade NLP classifiers, including spam filtering, fake news detection,
and sentiment analysis. The greedy attack finds semantically similar
words (enumerating all possibilities to find words with a minimal
distance and score difference from the original input) and replacing
them in sentences with a high language model score. Three classifiers
were evaded: NB, LSTM, and a word-level 1D CNN.

\citet{DBLP:journals/corr/abs-1812-00151} did the same while using
a joint sentence and word paraphrasing technique to maintain the original
semantics and syntax of the text. They attacked LSTM and a word-level
1D CNN trained on same datasets used by \citet{kuleshov2018adversarial},
providing a more effective attack on many datasets, including a spam
filtering dataset.

This interesting approach of generalization can be extended in the
future by applying other NLP-based attacks in the domain of spam adversarial
examples.

\subsection{\label{subsec:Cyber-Physical-Systems-(CPS)}Cyber-Physical Systems
(CPSs) and Industrial Control Systems (ICSs)}

Cyber-physical systems (CPSs) and industrial control systems (ICSs)
consist of hardware and software components that control and monitor
physical processes, e.g., critical infrastructure, including the electric
power grid, transportation networks, water supply networks, nuclear
plants, and autonomous car driving. A summary of the attacks in the
CPS sub-sdomain is provided in Table \ref{tab:Comparison-of-Adversarial-5}. 

\begin{table}
\caption{\label{tab:Comparison-of-Adversarial-5}Summary of Adversarial Learning
Approaches in Cyber-Physical Systems}

\centering{}%
\begin{tabular}{>{\centering}p{0.09\textwidth}>{\centering}p{0.04\textwidth}>{\centering}p{0.15\textwidth}>{\centering}p{0.08\textwidth}>{\centering}p{0.1\textwidth}>{\centering}p{0.05\textwidth}>{\centering}p{0.1\textwidth}>{\centering}p{0.03\textwidth}>{\centering}p{0.03\textwidth}>{\centering}p{0.03\textwidth}>{\centering}p{0.03\textwidth}>{\centering}p{0.03\textwidth}}
Citation & Year & Target Classifier & Attack Type & Attack's Output & Threat Model & Perturbed Features & \rotatebox{90}{Reasonable attacker knowledge?} & \rotatebox{90}{End-to-end attack?} & \rotatebox{90}{Effective attack?} & \rotatebox{90}{Representative dataset?} & \rotatebox{90}{Representative features?}\tabularnewline
\midrule
\citet{DBLP:conf/indin/SpechtONH18} & 2018 & FC DNN & Inference integrity & Feature vector & White-box & Sensor signals & - & - & + & + & +\tabularnewline
\midrule
\citet{DBLP:conf/ijcai/GhafouriVK18} & 2018 & LR, DNN & Inference integrity & Feature vector & Gray-box & Sensor data & + & - & + & + & +\tabularnewline
\midrule
\citet{Clark2018AMA} & 2018 & RL Q-Learning & Inference integrity & Feature vector & White-box & Ultrasonic collision avoidance sensor data & - & - & + & + & +\tabularnewline
\midrule
\citet{Feng2017ADL} & 2017 & LSTM & Inference integrity & Feature vector & Gray-box & Sensor data & + & - & + & + & +\tabularnewline
\midrule
\citet{DBLP:journals/corr/abs-1907-07487} & 2019 & Autoencoders & Inference integrity/ availability & Feature vector & Gray-box  & Sensor data & + & - & + & +  & +\tabularnewline
\midrule
\citet{Yaghoubi:2019:GAT:3302504.3311814} & 2019 & RNN & Inference integrity & Feature vector & White-box & Continuous sensor data & - & - & + & + & +\tabularnewline
\midrule
\citet{li2020conaml} & 2020 & FC DNN & Inference integrity & Feature vector & Gray-box & Sensor data & + & - & + & + & +\tabularnewline
\bottomrule
\end{tabular}
\end{table}

\citet{DBLP:conf/indin/SpechtONH18} trained a fully connected DNN
on the SECOM dataset, recorded from a semi-conductor manufacturing
process, which consists of 590 attributes collected from sensor signals
and variables during manufacturing cycles. Each sensor data entry
is labeled as either a normal or anomalous production cycle. They
used the FGSM white-box inference integrity feature vector attack
to camouflage abnormal/dangerous sensor data so it appears normal.
This attack uses a white-box approach, making it less feasible in
real-life scenarios.

\citet{DBLP:conf/ijcai/GhafouriVK18} conducted a gray-box inference
integrity attack on a linear regression-based anomaly detector, a
neural network regression anomaly detector, and an ensemble of both,
using the TE-PCS dataset, which contains sensor data describing two
simultaneous gas-liquid exothermic reactions for producing two liquid
products. There are safety constraints that must not be violated (e.g.,
safety limits for the pressure and temperature of the reactor), corresponding
to the data. For the linear regression-based anomaly detector, the
problem of finding adversarial examples of sensor data can be solved
using a mixed-integer linear programming (MILP) problem. In order
to bypass the neural network regression and ensemble, an iterative
algorithm is used. It takes small steps in the direction of increasing
objective function. In each iteration, the algorithm linearizes all
of the neural networks at their operating points and solves the problem
using MILP as before. The mathematical approach taken in this attack
(MILP) can be used in the future to attack different models, e.g.,
auto-regressive models, with greater efficiency than current approaches.

\citet{Clark2018AMA} used a Jaguar autonomous vehicle (JAV) to emulate
the operation of an autonomous vehicle. The driving and interaction
with the JAV environment used the Q-learning reinforcement learning
algorithm. JAV acts as an autonomous delivery vehicle and transports
deliveries from an origination point to a destination location. The
attacker\textquoteright s goal is to cause the JAV to deviate from
its learned route to an alternate location where it can be ambushed.
A white-box inference integrity attack was chosen with the goal of
compromising the JAV\textquoteright s reinforcement learning (RL)
policy and causing the deviation. The attack was conducted by inserting
an adversarial data feed into the JAV via its ultrasonic collision
avoidance sensor. This attack uses a white-box approach, making it
less feasible in real-life scenarios. However, it attacks reinforcement
learning models which are frequently used in CPSs.

\citet{Feng2017ADL} presented a gray-box inference integrity attack
against a LSTM anomaly detector using a GAN (see Appendix \ref{sec:Appendix-A:-Deep})
with a substitute model as a discriminator. Two use cases were evaluated:
gas pipeline and water treatment plant sensor data. \citet{li2020conaml}
presented a gray-box inference integrity attack against the same dataset,
but used a constraint-based adversarial machine learning to adhere
the intrinsic constraints of the physical systems, modeled by mathematical
constraints derived from normal sensor data. This attack targets LSTM
classifiers, which are commonly used in ICS, making this an important
attack.

\citet{DBLP:journals/corr/abs-1907-07487} demonstrated inference
integrity and availability attacks against an autoencoder-based anomaly
detection system of water treatment plant sensor data. Access to both
the ICS features and benign sensor data (to train a substitute model)
is assumed, making this attack a gray-box attack. This attack enumerates
all possible operations for every sensor. This attack targets autoencoder
classifiers, which are commonly used to detect anomalies, as done
in ICSs, making this an important attack.

\citet{Yaghoubi:2019:GAT:3302504.3311814} presented a gray-box inference
integrity attack against a steam condenser with an RNN-based anomaly
detector that monitors continuous (e.g., signal) data. The attack
uses gradient-based local search with either uniform random sampling
or simulated annealing optimization to find the data to modify.

\subsection{Biometric Systems}

In this subsection, we focus on attacks that target the most commonly
used biometric systems that leverage machine learning: face, speech,
and iris recognition systems. While these attacks can be viewed as
computer vision (or audio) attacks, we focus only on such attacks
that affect cyber security authentication thus allowing unauthorized
users access to the protected systems. Many studies have focused on
adversarial attacks against other types of biometric systems, for
instance: handwritten signature verification (\citet{Hafemann2019CharacterizingAE}),
EEG biometrics (\citet{zdenizci2019AdversarialDL}), and gait biometric
recognition (\citet{Prabhu2017VulnerabilityOD}). However, as previously
mentioned, these are not discussed here due to space limitations.
A summary of the attacks against biometric systems is presented in
Table \ref{tab:Comparison-of-Adversarial-6}. 

\begin{table}
\caption{\label{tab:Comparison-of-Adversarial-6}Summary of Adversarial Learning
Approaches in Biometric Systems}

\centering{}%
\begin{tabular}{>{\centering}p{0.1\textwidth}>{\centering}p{0.04\textwidth}>{\centering}p{0.1\textwidth}>{\centering}p{0.1\textwidth}>{\centering}p{0.08\textwidth}>{\centering}p{0.1\textwidth}>{\centering}p{0.1\textwidth}>{\centering}p{0.07\textwidth}>{\centering}p{0.03\textwidth}>{\centering}p{0.03\textwidth}>{\centering}p{0.03\textwidth}>{\centering}p{0.03\textwidth}>{\centering}p{0.03\textwidth}}
Citation & Year & Target Classifier & Biometric Application & Attack Type & Attack's Output & Threat Model & Perturbed Features & \rotatebox{90}{Reasonable attacker knowledge?} & \rotatebox{90}{End-to-end attack?} & \rotatebox{90}{Effective attack?} & \rotatebox{90}{Representative dataset?} & \rotatebox{90}{Representative features?}\tabularnewline
\midrule
\citet{Sharif:2016:ACR:2976749.2978392} & 2016 & CNN & Face recognition & Inference integrity & Physical eyeglasses (end-to-end) & Feature-targeted white-box and black-box & Image's pixels & + & + & + & + & +\tabularnewline
\midrule
\citet{Trojannn} & 2018 & CNN & Face recognition & Training integrity & Image (non-physical end-to-end) & Feature-targeted white-box; \emph{add} training set access & Image's pixels & - & + & + & + & +\tabularnewline
\midrule
\citet{DBLP:journals/corr/abs-1712-05526} & 2017 & CNN & Face recognition & Training integrity & Physical accessory (end-to-end) & Feature-targeted white-box; \emph{add} training set access & Image's pixels & - & + & + & + & +\tabularnewline
\midrule
\citet{Kreuk2018FoolingES}  & 2018  & LSTM/GRU & Speaker recognition & Inference integrity & Feature vector & White-box or black-box & Mel-spectrum features and MFCCs & + & - & + & + & -\tabularnewline
\midrule
\citet{Gong2017CraftingAE} & 2017 & Mixed CNN-LSTM & Speaker recognition & Inference integrity & Raw waveform (end-to-end) & Black-box & Raw waveforms & + & + & + & + & +\tabularnewline
\midrule
\citet{Du2019SirenAttackGA} & 2019 & CNN & Speaker recognition & Inference integrity & Raw waveform (end-to-end) & Black-box & Raw waveforms & + & + & + & + & +\tabularnewline
\midrule
\citet{Cai2018AttackingSR} & 2018 & CNN & Speaker recognition & Inference integrity & Feature vector & Gray-box & Mel-spectrum features & + & - & + & + & -\tabularnewline
\midrule
\citet{Wang18-217482} & 2018 & CNN & Face recognition, Iris recognition & Inference integrity & Image (non-physical end-to-end) & Gray-box & Image's pixels & + & + & + & + & +\tabularnewline
\midrule
\citet{Taheri19} & 2019 & CNN & Fingerprint recognition, iris recognition & Inference integrity & Image (non-physical end-to-end) & White-box & Image's pixels & - & + & + & + & +\tabularnewline
\midrule
\citet{Soleymani19,Soleymani2019AdversarialET} & 2019 & CNN & Iris recognition & Inference integrity & Feature vector & Gray-box & Iris codes & + & - & + & + & -\tabularnewline
\bottomrule
\end{tabular}
\end{table}

\subsubsection{\label{subsec:Face-Recognition}Face Recognition}

\citet{Sharif:2016:ACR:2976749.2978392} presented an inference integrity
attack against face recognition systems. The target classifier for
the white-box attack was VGG-Face, a 39 layer CNN (\citet{Parkhi15}).
The target classifier for the black-box attack was the \href{https://www.faceplusplus.com/}{Face++ cloud service}.
Instead of perturbing arbitrary pixels in the facial image, this attack
only perturbed the pixels of eyeglasses which were worn by the attacker
so the attacker would either be classified as another person (label-target
attack) or not be classified as himself (indiscriminate attack). Both
attacks are feature-targeted. Sharif et al. used a commodity inkjet
printer to print the front plane of the eyeglass frames on glossy
paper, which they then affixed to the actual eyeglass frames when
physically performing attacks. This makes this end-to-end attack interesting,
as it can be used (in its black-box variant) in real-life scenarios.

Both \citet{DBLP:journals/corr/abs-1712-05526} and \citet{Trojannn}
used a black-box training integrity poisoning attack against face
recognition systems. The CNN target classifiers were VGG-Face (\citet{Parkhi15})
and DeepID (\citet{Sun:2014:DLF:2679600.2679769}) respectively. Liu
et al. used a non-physical image, such as a square appearing in the
picture, as the Trojan trigger in the picture to be labeled as a different
person. In \citet{DBLP:journals/corr/abs-1712-05526}, the poisoned
facial images contained a physical accessory as the key; a photo of
a person taken directly from the camera can then become a backdoor
when the physical accessory is worn; thus, this is both a feature-targeted
and a label-targeted attack. Both attacks require training set \emph{add}
access. Despite the fact that these are white-box attacks, the usage
of feature-targeting fits nicely with some forms of real-life attacks,
such as supply-chain attacks, where a powerful adversary wants to
exploit a very specific scenario (e.g., evade detection of a person
only when he/she is holding a key).

\subsubsection{\label{subsec:Speaker-Verification/Recognition}Speaker Verification/Recognition}

Note that in this subsection, we only discuss speaker recognition
and verification system adversarial attacks.

\citet{Kreuk2018FoolingES} presented white-box inference integrity
attacks on an LSTM/GRU classifier that was either trained on the YOHO
or NTIMIT datasets using two types of features: Mel-spectrum features
and MFCCs. They also presented two black-box inference integrity attacks,
using the transferability property. In the first one, they generated
adversarial examples with a system trained on NTIMIT dataset and performed
the attack on a system that was trained on YOHO. In the second attack,
they generated the adversarial examples with a system trained using
Mel-spectrum features and performed the attack on a system trained
using MFCCs. All of the attacks used the FGSM attack, and the attack
output was a feature vector and not a complete audio sample. This,
and the fact that this attack uses a white-box approach, making it
less feasible in real-life scenarios

\citet{Gong2017CraftingAE} trained a WaveRNN model (a mixed CNN-LSTM
model) on raw waveforms (the IEMOCAP dataset's utterances; see Table
\ref{tab:Commonly-Used-Datasets}) for speaker recognition, as well
as emotion and gender recognition. They used a substitute waveCNN
model and performed a black-box inference integrity attack using FGSM
on the raw waveforms, rather than on the acoustic features, making
this an end-to-end attack that does not require an audio reconstruction
step. The use of a gray-box approach and raw waveform features makes
this attack feasible in real-life scenarios.

\citet{Du2019SirenAttackGA} used six state-of-the-art speech command
recognition CNN models: VGG19, DenseNet, ResNet18, ResNeXt, WideResNet18,
and DPN-92, all adapted to the raw waveform input. The models were
trained for speaker recognition on the IEMOCAP dataset (see Table
\ref{tab:Commonly-Used-Datasets}) and for speech recognition, sound
event classification, and music genre classification using different
datasets. The black-box inference integrity attack used FGSM or particle
swarm optimization (PSO) on the raw waveforms. The use of a gray-box
approach and raw waveform features and the evaluation against many
classifiers make this attack feasible in real-life scenarios.

\citet{Cai2018AttackingSR} trained a CNN classifier that performs
multi-speaker classification using Mel-spectrograms as input. They
used a Wasserstein GAN with gradient penalty (WGAN-GP) to generate
adversarial examples for an indiscriminate gray-box inference integrity
attack and also used a WGAN-GP with a modified objective function
for a specific speaker for a targeted attack. The attack output is
a feature vector of Mel-spectrograms and not an audio sample. This
attack uses a white-box approach, making it less feasible in real-life
scenarios.

\subsubsection{Iris and Fingerprint Systems}

\citet{Wang18-217482} performed an indiscriminate black-box inference
integrity attack, leveraging the fact that many image-based models,
including face recognition and iris recognition models, use transfer
learning, i.e., they add new layers on top of pretrained layers which
are trained on a different model (a teacher model with a known architecture)
and are used to extract high-level feature abstractions from the raw
pixels. For instance, the face recognition model's teacher model can
be VGG-Face (\citet{Parkhi15}), while an iris model's teacher model
can be VGG16. By attacking the teacher model using white-box attacks,
such as C\&W, the target model (student model), for which the architecture
is unknown, is also affected. This approach can be used to attack
many classifiers, especially those in domains where transfer learning
using a known model is common, like the computer vision domain.

\citet{Taheri19} trained a CNN classifier on the CASIA dataset of
images of iris and fingerprint data. They implemented a white-box
inference integrity attack using the FGSM, JSMA, DeepFool, C\&W, and
PGD methods to generate the perturbed images. This attack uses a white-box
approach, making it less feasible in real-life scenarios.

\citet{Soleymani19,Soleymani2019AdversarialET} generated adversarial
examples for code-based iris recognition systems, using a gray-box
inference integrity attack. However, conventional iris code generation
algorithms are not differentiable with respect to the input image.
Generating adversarial examples requires backpropagation of the adversarial
loss. Therefore, they used a deep autoencoder substitute model to
generate iris codes that were similar to iris codes generated by a
conventional algorithm (OSIRIS). This substitute model was used to
generate the adversarial examples using FGSM, iGSM, and Deepfool attacks.
The attack of iris codes can serve as the initial step toward end-to-end
adversarial attacks against biometric systems.

\textcolor{red}{}%

\section{\label{sec:Adversarial-Defense-Methods}Adversarial Defense Methods
in the Cyber Security Domain}

Our taxonomy focuses on the attack side, but every attack is accompanied
by a corresponding defense method.

If adversarial attacks are equivalent to malware attacking a computer
(a machine learning model), then defense methods can be viewed as
anti-malware products. However, most defense methods have been evaluated
in the image recognition domain for CNNs and in the NLP domain for
RNNs. Due to space limitations, we cannot provide a complete list
of the state-of-the-art prior work in those domains. A more comprehensive
list can be found, e.g., in \citet{Qiu19}.

Several papers presenting attacks in the cyber security domain (e.g.,
\citet{Grosse2017,DBLP:journals/corr/abs-1902-08909}) discuss the
fact that the attack is effective even in the presence of well-known
defense methods that were evaluated and found effective in the computer
vision domain (e.g., distillation and adversarial retraining). However,
only a few defense methods were developed specifically for the cyber
security domain and its unique challenges, like those described in
Section \ref{sec:Preliminary-Discussion:-The}. Furthermore, cyber
security classifiers usually have a different architecture than computer
vision classifiers, against which most of the methods presented in
prior research were evaluated.

Adversarial defense methods can be divided into two subgroups: (1)
detection methods - methods used to detect adversarial examples, and
(2) robustness methods - methods used to make a classifier more robust
to adversarial examples, without explicitly trying to detect them.

Each defense method is either \emph{attack-specific} (i.e., it requires
adversarial examples generated by the attack in order to mitigate
the attack) or \emph{attack-agnostic} (i.e., it works against all
types of attack methods, without the need to have a dataset of adversarial
examples generated by those attacks). Attack-agnostic defense methods
are more generic and are therefore preferable.

In the malware detection sub-domain, \citet{CHEN2018326} suggested
an attack-agnostic method to make an Android malware classifier robust
to poisoning attacks. Their method includes two phases: an offline
training phase that selects and extracts features from the training
set and an online detection phase that utilizes the classifier trained
by the first phase. These two phases are intertwined through a self-adaptive
learning scheme, in which an automated camouflage detector is introduced
to filter the suspicious false negatives and feed them back into the
training phase. \citet{DBLP:journals/corr/abs-1712-05919} evaluated
three attack-agnostic robustness defense methods: weight decay, an
ensemble of classifiers, and distillation (neural network compression,
resulting in gradient masking) for a dynamic analysis malware classifier
based on a non-sequence based deep neural network. \citet{DBLP:journals/corr/abs-1901-09963}
tried to defend an API call-based RNN classifier and compared their
own RNN attack-agnostic robustness defense method, termed sequence
squeezing, to four robustness defense methods inspired by existing
CNN-based defense methods: adversarial retraining (retraining the
classifier after adding adversarial examples), defense GAN (generating
several samples using a GAN and using the closest one to the original
input as the classifier's input), nearest neighbor classification,
and RNN ensembles. They showed that sequence squeezing provides the
best trade-off between training and inference overhead (which is less
critical in the computer vision domain) and adversarial robustness.

For DGA detection, Sidi et al. \citep{DBLP:journals/corr/abs-1902-08909}
evaluated the robustness defense methods of adversarial retraining
(attack-specific) and distillation (attack-agnostic), showing that
they improve the robustness against adversarial attacks.

For CPS defense, Kravchic and Shabtai \citet{DBLP:journals/corr/abs-1907-01216}
suggested detecting adversarial attacks in ICS data using 1D CNNs
and undercomplete autoencoders (UAEs), an attack-agnostic method.
The authors demonstrate that attackers must sacrifice much of their
attack's potential damage to remain hidden from the detectors. \citet{DBLP:conf/ijcai/GhafouriVK18}
presented robust linear regression and neural network regression-based
anomaly detectors for CPS anomalous data detection by modeling a game
between the defender and attacker as a Stackelberg game in which the
defender first commits to a collection of thresholds for the anomaly
detectors, and the attacker then computes an optimal attack. The defender
aims to minimize the impact of attacks, subject to a constraint typically
set to achieve a target number of false alarms without consideration
of attacks.

For spam detection, \citet{DBLP:conf/emnlp/AlzantotSEHSC18} evaluated
the attack-specific robustness defense methods of adversarial retraining
against RNN classifiers, showing that such methods improve robustness
against adversarial attacks.

For biometric systems defense methods, \citet{Taheri19} presented
an attack-specific detection architecture that includes shallow and
deep neural networks to defend against biometric adversarial examples.
The shallow neural network is responsible for data preprocessing and
generating adversarial samples. The deep neural network is responsible
for understanding data and information, as well as for detecting adversarial
samples. The deep neural network gets its weights from transfer learning,
adversarial training, and noise training. \citet{DBLP:conf/indin/SpechtONH18}
suggested an attack-specific robustness defense method using an iterative
adversarial retraining process to mitigate adversarial examples for
semi-conductor anomaly detection system of sensor data. \citet{Soleymani19}
used an attack-agnostic robustness defense method involving wavelet
domain denoising of the iris samples, by investigating each wavelet
sub-band and removing the sub-bands that are most affected by the
adversary.

In our opinion, other defense methods proposed for the computer vision
domain could inspire similar defense methods in the cyber domain.
However, their effectiveness in the cyber security domain would need
to be evaluated, since as discussed above, the cyber security domain
is different and has unique challenges. Furthermore, in the cyber
security domain further emphasis should be put on the defense method
overhead (as done, e.g., in \citet{DBLP:journals/corr/abs-1901-09963}),
since cyber security classifiers usually perform their classification
in real-time, meaning that low overhead is critical, unlike in the
computer vision domain. Finally, we believe that te research attention
should be given to attack-agnostic defense methods, as attack-specific
defense methods (e.g., adversarial retraining) provide a very narrow
defense against the growing variety of adversarial attacks presented
in this study.

\section{\label{sec:Adversarial-Learning-FutureUniqu}Current Gaps and Future
Research Directions for Adversarial Learning in the Cyber Security
Domain}

In this section, we highlight gaps in our taxonomy, presented in Section
\ref{sec:Taxonomy}, which are not covered by the applications presented
in Sections \ref{sec:Cyber-Applications-of} and \ref{sec:Adversarial-Defense-Methods},
despite having a required functionality. Each such gap is presented
in a separate subsection below. For each gap, we summarize the progress
made on this topic in other domains of adversarial learning, such
as the computer vision domain, and extrapolate future trends in the
cyber security domain from it.

\subsection{\label{subsec:Feature-Targeted-Attacks}Attack's Targeting Gap: Feature-Targeted
Attacks and Defenses}

Poisoning integrity attacks place mislabeled training points in a
region of the feature space far from the rest of the training data.
The learning algorithm labels such a region as desired, allowing for
subsequent misclassifications at test time. However, adding samples
to the training set might cause misclassification of many samples
and thus would raise suspicion of an attack, while the adversary might
want to evade just a specific sample (e.g., a dedicated APT).

In non-cyber security domains, a feature-targeted attack, also known
as a Trojan neural network attack (\citet{Trojannn}) or backdoor
attack (\citet{DBLP:journals/access/GuLDG19}, \citet{DBLP:journals/corr/abs-1712-05526}),
is a special form of poisoning attack, which aims to resolve this
problem. A model poisoned by a backdoor attack should still perform
well on most inputs (including inputs that the end user may hold out
as a validation set) but cause targeted misclassifications or degrade
the accuracy of the model for inputs that satisfy some secret, attacker-chosen
property, which is referred to as the backdoor trigger. For instance,
adding a small rectangle to a picture would cause it to be classified
with a specific target label \citet{Trojannn}. Such attacks were
performed on face recognition \citet{DBLP:journals/corr/abs-1712-05526,Trojannn},
traffic sign detection (\citet{DBLP:journals/access/GuLDG19}), sentiment
analysis, speech recognition, and autonomous driving (\citet{Trojannn})
datasets. 

However, such attacks have not been applied yet in the cyber security
domain, despite the fact that such attacks have interesting use cases.
For instance, such an attack might allow only a specific nation-state
APT to bypass the malware classifier, while still detecting other
malware, leaving the defender unaware of this feature-targeted attack.

Defenses against such attacks are also required. In the image recognition
domain, \citet{DBLP:conf/sp/WangYSLVZZ19} generates a robust classifier
by pruning backdoor-related neurons from the original DNN. \citet{Gao19}
detects a Trojan attack during run-time by perturbing the input and
observing the randomness of predicted classes for perturbed inputs
on a given deployed model. A low entropy in predicted classes implies
the presence of a Trojaned input. Once feature-targeted attacks are
published in the cyber security domain, defense methods to mitigate
them will follow.

\subsection{\label{subsec:Attacker's-Goals-Gap:}Attacker's Goals Gap: Resource
Exhaustion-Based Availability Attacks}

There are two availability attack variants. One is the subversion
attack, in which the system is fooled so it does not serve legitimate
users. Such adversarial attacks have been implemented in the cyber
security domain by fooling the classifier into considering legitimate
traffic as malicious, thus blocking it (e.g., \citet{DBLP:conf/ccs/HuangJNRT11,DBLP:conf/nsdi/NelsonBCJRSSTX08}).
The other variant is the resource exhaustion attack, in which all
of a system\textquoteright s available resources are used in order
to prevent legitimate users from using them. The second variant is
very common in the cyber security domain (e.g., distributed denial-of-service
attacks on websites, zip bombs, etc.) but not in adversarial attacks
in this domain.

Given recent advancements in the computer vision domain, e.g., sponge
examples \citet{shumailov2020sponge}, which are adversarial examples
whose classification requires 10-200 times the resources of classifying
a regular sample, we believe it is only a matter of time before such
examples will reach the cyber security domain, e.g., PE files that
take a very long to classify.

\subsection{Attacker's Goals and Knowledge Gap: Confidentiality Attacks via Model
Queries and Side-Channels}

Reverse engineering (reversing) of traditional (non-ML-based) anti-malware
is a fundamental part of a malware developer\textquoteright s modus
operandi. So far, confidentiality attacks have only been conducted
against image recognition models and not against cyber security related
models, e.g., malware classifiers. However, performing such attacks
in the cyber security domain might provide the attacker with enough
data to perform more effective white-box attacks, instead of black-box
ones.

In the image recognition domain, confidentiality attacks have been
conducted by querying the target model. \citet{DBLP:conf/uss/TramerZJRR16}
formed a query-efficient gray-box (the classifier type should be known)
score-based attack. The attack used equation solving to recover the
model's weights from sets of observed sample-confidence score pairs
$(x,h(x))$ retrieved by querying the target model. For instance,
a set of $n$ such pairs is enough to retrieve the $n$ weights of
a logistic regression classifier using $n$-dimensional input vectors.
\citet{Wang2018} used a similar approach to retrieve the model's
hyperparameters (e.g., the factor of the regularization term in the
loss function equation). 

In non-cyber domains, confidentiality attacks have also been conducted
via side-channel information. \citet{DBLP:journals/corr/abs-1812-11720}
used timing attack side-channels to obtain neural network architecture
information. \citet{Batina19} used electromagnetic radiation to get
neural network architecture information from embedded devices. \citet{Hua:2018:REC:3195970.3196105}
used both a timing side-channel and off-chip memory access attempts
during inference to discover on-chip CNN model architecture.

\subsection{Perturbed Features Gap: Exploiting Vulnerabilities in Machine and
Deep Learning Frameworks}

In non-ML based cyber security solutions, vulnerabilities are commonly
used to help attackers reach their goals (e.g., use buffer overflows
to run adversary-crafted code in a vulnerable process). A vulnerability
in the underlying application or operating system is a common attack
vector for disabling or bypassing a security product. This trend is
starting to be seen in the adversarial learning domain against image
classifiers, however such attacks have not been demonstrated against
malware classifiers. Such vulnerabilities are specialized, and should
be researched explicitly for the proper cyber use cases in order to
be effective.

In the image recognition domain, \citet{Xiao18} discovered security
vulnerabilities in popular deep learning frameworks (Caffe, TensorFlow,
and Torch). \citet{DBLP:journals/corr/StevensSRHHD17} did the same
for OpenCV (a computer vision framework), scikit-learn (a machine
learning framework), and Malheur \citet{DBLP:journals/jcs/RieckTWH11}
(a dynamic analysis framework used to classify unknown malware using
a clustering algorithm). By exploiting these frameworks' implementations,
attackers can launch denial-of-service attacks that crash or hang
a deep learning application, or control-flow hijacking attacks that
lead to either system compromise or recognition evasions. 

We believe that future adversarial attacks can view the deep learning
framework used to train and run the model as a new part of the attack
surface, leveraging vulnerabilities in the framework, which can even
be detected automatically by a machine learning model (as reviewed
in \citet{Ghaffarian:2017:SVA:3135069.3092566}). Some of these vulnerabilities
can be used to add data to input in a way that would cause the input
to be misclassified, just as adversarial perturbation would, but by
subverting the framework instead of subverting the algorithm. This
can be viewed as an extension of the perturbed features attack characteristics
in our taxonomy.

\subsection{Attack's Output Gap: End-to-End Attacks in Complex Format Sub Domains}

As discussed in Section \ref{subsec:End-to-end-Attacks}, only end-to-end
attacks can be used to attack systems in the cyber security domain.
Some sub-domains, such as email, have a simple format and therefore
it is easier to map from features (words) back to a sample (email)
and create an end-to-end attack. However, in more complex sub-domains,
such as NIDS and CPS, the mapping from features to a full sample (e.g.,
a network stream) is complex. As can be seen in Sections \ref{subsec:Network-Intrusion-Detection}
and \ref{subsec:Cyber-Physical-Systems-(CPS)}, only a small number
of attacks in the NIDS and CPS sub-domains (less than 10\%) are end-to-end
attacks.

We predict that the trend seen in the computer vision domain, where
after several years of feature vector attacks (e.g., \citet{Goodfellow14}),
end-to-end attacks followed \citet{roadsigns17}, will also be seen
in the cyber security domain. There will likely be three directions
for such end-to-end attacks: 1) Adding new features to an existing
sample, e.g., \citet{li2020feature,DBLP:conf/raid/RosenbergSRE18,DBLP:conf/sp/SrndicL14}.
2) Modifying only a subset of features that can be modified without
harming the functionality of an existing sample, e.g., \citet{RosenbergIJCNN2020,Kuppa:2019:BBA:3339252.3339266}.
3) Using cross-sample transformations (e.g., packing) that would change
many features simultaneously \citet{RosenbergIJCNN2020,DBLP:journals/corr/abs-1801-08917}.

\subsection{\label{subsec:Adversarial-Defense-Methods}Adversarial Defense Method
Gaps}

The gaps in the domain of defense methods against adversarial attacks
in the cyber security domain is acute, because this domain involves
actual adversaries: malware developers who want to evade next generation
machine and deep learning-based classifiers. Such attacks have already
been executed in the wild against static analysis deep neural networks
\citet{CylanceAdvAttack}. We have mapped two gaps specific to the
cyber security domain, which are shown below.

\subsubsection{\label{subsec:Evaluating-The-Robustness}Metrics to Measure the Robustness
of Classifiers to Adversarial Examples}

Several papers (\citet{10.1007/978-3-319-63387-9_5,weng2018evaluating,Peck:2017:LBR:3294771.3294848})
suggested measuring the robustness of machine learning systems to
adversarial attacks by approximating the lower bound on the perturbation
needed for any adversarial attack to succeed; the larger the perturbation,
the more robust the classifier. However, these papers assume that
the robustness to adversarial attacks can be evaluated by the minimal
perturbation required to modify the classifier's decision. This raises
the question of whether this metric is valid in the cyber security
domain.

Section \ref{subsec:Small-Perturbations-Are} leads us to the conclusion
that \emph{minimal perturbation is not necessarily the right approach
for adversarial learning in the cyber security domain}. As already
mentioned in \citet{DBLP:conf/ccs/BiggioR18}, maximum confidence
attacks, such as the Carlini and Wagner (C\&W) attack (Section \ref{sec:Adversarial-Learning-Methods}),
are more effective. However, this is not the complete picture.

As mentioned in Section \ref{subsec:Executables-are-More}, in the
cyber security domain, classifiers usually use more than a single
feature type as input (e.g., both PE header metadata and byte entropy
in \citet{Saxe2015}). Certain feature types are easier to modify
without harming the executable functionality than others. On the other
hand, an attacker can add as many strings as needed; in contrast to
images, adding more strings (i.e., a larger perturbation) is not more
visible to the user than adding less strings, since the executable
file is still a binary file.

This means that we should not only take into account the impact of
a feature on the prediction, but also the difficulty of modifying
this feature type. Unfortunately, there is currently no numeric metric
to assess the difficulty of modifying features. Currently, we must
rely on the subjective opinion of experts who assess the difficulty
of modifying each feature type, as shown in Katzir and Elovici \citet{KATZIR2018419}.
When such a metric becomes available, combining it with the maximum
impact metric would be a better optimization constraint than minimal
perturbation.

In conclusion, both from an adversary's perspective (when trying to
decide which attack to use) and from the defender's perspective (when
trying to decide which classifier would be the most robust to adversarial
attack), the metric of evaluation currently remains an open question
in the cyber security domain.

\subsubsection{Defense Methods Combining Domain-Specific and Domain-Agnostic Techniques}

Domain-specific constraints and properties have been leveraged in
the computer vision domain. For instance, the \emph{feature squeezing}
method mentioned in \citet{DBLP:conf/ndss/Xu0Q18} applied image-specific
dimensionality reduction transformations to the input features, such
as changing the image color depth (e.g., from 24 bit to 8 bit), in
order to detect adversarial examples in images.

We argue that the same approach can be applied in the cyber security
domain, as well. In the cyber security domain, there are very complex
constraints on the raw input. For instance, a PE file has a very strict
format (in order to be loadable by the Windows operating system),
with severe limitations on the PE header metadata values, etc. Such
constraints can be leveraged to detect ``domain-specific anomalies''
more accurately. For example, the attacks used to evade malware classifiers
in \citet{RosenbergBHEU20,RosenbergIJCNN2020} involved concatenation
of strings from the IAT to the EOF/PE overlay; while such concatenation
generates a valid PE, one might ask whether it makes sense for such
strings to be appended to the EOF instead of, e.g., being in the data
section or a resource section.

\subsubsection{Defense Methods Robust to Unknown and Transparent-Box Adversarial
Attacks}

There are two main challenges when developing a defense method:

The first challenge is creating a defense method which is also robust
against transparent-box attacks, i.e., attackers who know what defense
methods are being used and select their attack methods accordingly.

In the computer vision domain, \citet{tramer2020adaptive,Carlini2017}
showed that many different types of commonly used defense methods
(e.g., detection of adversarial examples using statistical irregularities)
are rendered useless by a specific type of adversarial attack. \citet{DBLP:conf/woot/HeWCCS17}
showed the same for feature squeezing, and \citet{DBLP:conf/icml/AthalyeC018,DBLP:conf/ccs/HashemiCK18,DBLP:journals/corr/abs-1712-09196}
presented similar results against Defense-GAN.

Similar research should be conducted in the cyber security domain.
For instance, attackers can make their attack more robust against
RNN subsequence model ensembles presented in \citet{DBLP:journals/corr/abs-1901-09963}
by adding perturbations across the entire API call sequence and not
just until the classification changes.

The second challenge is creating defense methods that are effective
against all attacks and not just specific ones, termed \emph{attack-agnostic}
defense methods in \citet{DBLP:journals/corr/abs-1901-09963}. However,
the challenge of finding a metric to evaluate the robustness of classifiers
to adversarial attacks in the cyber security domain, already discussed
in Section \ref{subsec:Evaluating-The-Robustness}, makes the implementation
of attack-agnostic defense methods in the cyber security domain more
challenging than in other domains.

\section{Conclusions}

In this paper, we reviewed the latest research on a wide range of
adversarial learning attacks and defenses in the cyber security domain
(e.g., malware detection, network intrusion detection, etc.). 

One conclusion based on our review is that while feature vector adversarial
attacks in the cyber security domain are possible, real-life attacks
(e.g., against next generation antivirus software) are challenging.
This is due to the unique challenges that attackers and defenders
are faced with in the cyber security domain; these challenges include
the difficulty of modifying samples end-to-end without damaging the
malicious business logic and the need to modify many feature types
with various levels of modification difficulty.

From the gaps in our taxonomy discussed above and from the recent
advancements in other domains of adversarial learning, we identified
some directions for future research in adversarial learning in the
cyber security domain. One direction focuses on the implementation
of feature-triggered attacks that would work only if a certain trigger
exists, leaving the system's integrity unharmed in other cases, thus
making it harder to detect the attack. Another possible direction
is performing confidentiality attacks involving model reversing via
queries or side-channels. A third research direction aims at expanding
the attack surface of adversarial attacks to include the vulnerabilities
in the relevant machine learning framework and designing machine learning
models to detect and leverage them. From the defender's point of view,
more robust defense methods against adversarial attacks in the cyber
security domain should be the focus of future research.

A final conclusion derived from our review is that adversarial learning
in the cyber security domain is becoming more and more similar to
the cat and mouse game played in the traditional cyber security domain,
in which the attackers implement more sophisticated attacks to evade
the defenders and vice versa. A key takeaway is that defenders should
become more proactive in assessing their system's robustness to adversarial
attacks, the same way penetration testing is proactively used in the
traditional cyber security domain.

\bibliographystyle{ACM-Reference-Format}
\bibliography{thesis}

\newpage{}

\appendix

\section{\label{sec:Appendix-A:-Deep}Deep Learning Classifiers: Mathematical
and Technical Background}

\subsection{Deep Neural Networks (DNNs)}

Neural networks are a class of machine learning models made up of
layers of neurons (elementary computing units).

A neuron takes an n-dimensional feature vector $\mathbf{x}=\left[x_{1},x_{2}...x_{n}\right]$
from the input or the lower level neuron and outputs a numerical output
$\mathbf{y}=\left[y_{1},y_{2}...y_{m}\right]$, such that 

\begin{equation}
y_{j}=\phi(\sideset{}{_{i=1}^{n}}\sum w_{ji}x_{i}+b_{j})
\end{equation}
\\
to the neurons in higher layers or the output layer. For the neuron
j, $y_{j}$ is the output and $b_{j}$ is the bias term, while $w_{ji}$
are the elements of a layer\textquoteright s weight matrix. The function
$\phi$ is the nonlinear activation function, such as $sigmoid()$,
which determines the neuron\textquoteright s output. The activation
function introduces nonlinearities to the neural network model. Otherwise,
the network remains a linear transformation of its input signals.
Some of the success of DNNs is attributed to these multi-layers of
nonlinear correlations between features, which are not available in
popular traditional machine learning classifiers, such as SVM, which
has at most a single nonlinear layer using the kernel trick.

A group of $m$ neurons forms a hidden layer which outputs a feature
vector $\mathbf{y}$. Each hidden layer takes the previous layer\textquoteright s
output vector as the input feature vector and calculates a new feature
vector for the layer above it:

\begin{equation}
\boldsymbol{y}_{l}=\phi(\boldsymbol{W}_{l}\boldsymbol{y}_{l-1}+\boldsymbol{b}_{l})\label{eq:-7}
\end{equation}
\\
where $\boldsymbol{y}_{l}$, $\boldsymbol{W}_{l}$ and $\boldsymbol{b}_{l}$
are the output feature vector, the weight matrix, and the bias of
the l-th layer, respectively. Proceeding from the input layer, each
subsequent higher hidden layer automatically learns a more complex
and abstract feature representation which captures a higher level
structure.

\subsubsection{Convolutional Neural Networks (CNNs)}

CNNs are a type of DNN. Let $\boldsymbol{x_{i}}$ be the k-dimensional
vector corresponding to the i-th element in the sequence. A sequence
of length n (padded when necessary) is represented as: $\boldsymbol{x}[0:n-1]=\boldsymbol{x}[0]\perp\boldsymbol{x}[1]\perp\boldsymbol{x}[n-1]$,
where $\perp$ is the concatenation operator. In general, let $\boldsymbol{x}[i:i+j]$
refer to the concatenation of words $\boldsymbol{x}[i],\boldsymbol{x}[i+1],...,\boldsymbol{x}[i+j]$
. A convolution operation involves a filter $\boldsymbol{w}$, which
is applied to a window of h elements to produce a new feature. For
example, a feature $c_{i}$ is generated from a window of words $\boldsymbol{x}[i:i+h-1]$
by:

\begin{equation}
c_{i}=\phi(\boldsymbol{W}\boldsymbol{x}[i:i+h]+b)
\end{equation}
\\
where $b$ is the bias term and $\phi$ is the activation function.
This filter is applied to each possible window of elements in the
sequence $\left\{ \boldsymbol{x}[0:h-1],\boldsymbol{x}[1:h],...,\boldsymbol{x}[n-h:n-1]\right\} $
to produce a \emph{feature map}: $\boldsymbol{c}=\left[c_{0},c_{1},...,c_{n-h}\right]$.
We then apply a max over time pooling operation over the feature map
and take the maximum value: $\hat{c}=\max(\boldsymbol{c})$ as the
feature corresponding to this particular filter. The idea is to capture
the most important feature (the one with the highest value) for each
feature map. 

We described the process by which one feature is extracted from one
filter above. The CNN model uses multiple filters (with varying window
sizes) to obtain multiple features. These features form the penultimate
layer and are passed to a fully connected softmax layer whose output
is the probability distribution over labels.

CNNs have two main differences from fully connected DNNs: 
\begin{enumerate}
\item CNNs exploit spatial locality by enforcing a local connectivity pattern
between neurons of adjacent layers. The architecture thus ensures
that the learned ''filters'' produce the strongest response to a
spatially local input pattern. Stacking many such layers leads to
nonlinear ``filters'' that become increasingly ``global.'' This
allows the network to first create representations of small parts
of the input and assemble representations of larger areas from them.
\item In CNNs, each filter is replicated across the entire input. These
replicated units share the same parameterization (weight, vector,
and bias) and form a feature map. This means that all of the neurons
in a given convolutional layer respond to the same feature (within
their specific response field). Replicating units in this way allows
for features to be detected regardless of their position in the input,
thus constituting the property of translation invariance. This property
is important in both vision problems and with sequence input, such
as API call traces.
\end{enumerate}

\subsection{Recurrent Neural Networks (RNNs)}

A limitation of neural networks is that they accept a fixed sized
vector as input (e.g., an image) and produce a fixed sized vector
as output (e.g., probabilities of different classes). Recurrent neural
networks can use sequences of vectors in the input, output, or both.
In order to do that, the RNN has a hidden state vector, the context
of the sequence, which is combined with the current input to generate
the RNN's output. 

Given an input sequence $\left[\boldsymbol{x}_{1},\boldsymbol{x}_{2}...\boldsymbol{x}_{T}\right]$,
the RNN computes the hidden vector sequence$\left[\boldsymbol{h}_{1},\boldsymbol{h}_{2}...\boldsymbol{h}_{T}\right]$
and the output vector sequence $\left[\boldsymbol{y}_{1},\boldsymbol{y}_{2}...\boldsymbol{y}_{T}\right]$
by iterating the following equations from $t=1$ to $T$: 

\begin{equation}
\boldsymbol{h}_{t}=\mathcal{\mathcal{\phi}}(\boldsymbol{W}_{xh}\boldsymbol{x}_{t}+\boldsymbol{W}_{hh}\boldsymbol{x}_{t-1}+\boldsymbol{b}_{h})\label{eq:-5}
\end{equation}

\begin{equation}
\boldsymbol{y}_{t}=\boldsymbol{W}_{hy}\boldsymbol{h}_{t}+\boldsymbol{b}_{o}\label{eq:-6}
\end{equation}
\\
where the $\mathbf{W}$ terms denote weight matrices (e.g., $\mathbf{\boldsymbol{W}_{xh}}$
is the input hidden weight matrix), the $\boldsymbol{b}$ terms denote
bias vectors (e.g., $\boldsymbol{b}_{h}$ is the hidden bias vector),
and $\phi$ is usually an element-wise application of an activation
function. DNNs without a hidden state, as specified in Equation \ref{eq:-5},
reduce Equation \ref{eq:-6} to the private case of Equation \ref{eq:-7},
known as \emph{feedforward networks}.

\subsubsection{Long Short-Term Memory (LSTM)}

Standard RNNs suffer from both exploding and vanishing gradients.
Both problems are caused by the RNNs' iterative nature, in which the
gradient is essentially equal to the recurrent weight matrix raised
to a high power. These iterated matrix powers cause the gradient to
grow or shrink at a rate that is exponential in terms of the number
of time steps $T$. The vanishing gradient problem does not necessarily
cause the gradient to be small; the gradient\textquoteright s components
in directions that correspond to long-term dependencies might be small,
while the gradient\textquoteright s components in directions that
correspond to short-term dependencies is large. As a result, RNNs
can easily learn the short-term but not the long-term dependencies.
For instance, a conventional RNN might have problems predicting the
last word in: ``I grew up in France...I speak fluent \emph{French}''
if the gap between the sentences is large.

The LSTM architecture (\citet{Hochreiter1997}), which uses purpose-built
memory cells to store information, is better at finding and exploiting
long-range context than conventional RNNs. The LSTM\textquoteright s
main idea is that instead of computing $\boldsymbol{h}_{t}$ from
$\boldsymbol{h}_{t-1}$ directly with a matrix-vector product followed
by a nonlinear transformation (Equation \ref{eq:-5}), the LSTM directly
computes $\triangle\boldsymbol{h}_{t}$ , which is then added to $\boldsymbol{h}_{t-1}$
to obtain $\boldsymbol{h}_{t}$. This implies that the gradient of
the long-term dependencies cannot vanish. 

\subsubsection{Gated Recurrent Unit (GRU)}

Introduced in \citet{Cho2014}, the gated recurrent unit (GRU) is
an architecture that is similar to LSTM, but reduces the gating signals
from three (in the LSTM model: input, forget, and output) to two.
The two gates are referred to as an update gate and a reset gate.
Some research has shown that a GRU RNN is comparable to, or even outperforms,
LSTM in many cases, while using less training time. 

\subsubsection{Bidirectional Recurrent Neural Networks (BRNNs)}

One shortcoming of conventional RNNs is that they are only able to
make use of previous context. It is often the case that for malware
events the most informative part of a sequence occurs at the beginning
of the sequence and may be forgotten by standard recurrent models.
Bidirectional RNNs (\citet{Schuster1997}) overcome this issue by
processing the data in both directions with two separate hidden layers,
which are then fed forward to the same output layer. A BRNN computes
the forward hidden sequence $\boldsymbol{\overrightarrow{h}}_{t}$
, the backward hidden sequence $\boldsymbol{\overleftarrow{h}}_{t}$,
and the output sequence $\boldsymbol{y}_{t}$ by iterating the backward
layer from $t=T$ to $1$ and the forward layer from $t=1$ to $T$,
and subsequently updating the output layer. Combining BRNNs with LSTM
results in bidirectional LSTM (\citet{Gravesa}), which can access
long-range context in both input directions.

\subsection{Generative Adversarial Networks (GANs)}

A GAN is a combination of two deep neural networks: a classification
network (the discriminator) which classifies between real and fake
inputs and a generative network (the generator) that tries to generate
fake inputs that would be misclassified as genuine by the discriminator
(\citet{DBLP:conf/nips/GoodfellowPMXWOCB14}), eventually reaching
a Nash equilibrium. The end result is a discriminator which is more
robust against fake inputs.

GANs are only defined for real-valued data, while RNN classifiers
use discrete symbols. The discrete outputs from the generative model
make it difficult to pass the gradient update from the discriminative
model to the generative model.

Modeling the data generator as a stochastic policy in reinforcement
learning can be done to bypass the generator differentiation problem
(\citet{DBLP:conf/aaai/YuZWY17}). 

\subsection{Autoencoders (AEs)}

Autoencoders are widely used for unsupervised learning tasks, such
as learning deep representations or dimensionality reduction. Typically,
a traditional deep autoencoder consists of two components, the encoder
and the decoder. Let us denote the encoder\textquoteright s function
as $f_{\theta}:X\rightarrow H$, and denote the decoder\textquoteright s
function as $g_{\omega}:H\rightarrow X$, where $\theta,\omega$ are
parameter sets for each function, $X$ represents the data space,
and $H$ represents the feature (latent) space. The reconstruction
loss is:

\begin{equation}
L(\theta,\omega)=\frac{1}{N}||X-g_{\omega}\left(f_{\theta}(X)\right)||^{2}
\end{equation}
\\
where $L(\theta,\omega)$ represents the loss function for the reconstruction.

\subsection{Deep Autoencoding Gaussian Mixture Model (DAGMM) }

The DAGMM (\citet{DBLP:conf/iclr/ZongSMCLCC18}) uses two different
networks, a deep autoencoder and a Gaussian mixture model (GMM) based
estimator network to determine whether a sample is anomalous or not.

\subsection{AnoGAN}

AnoGAN (\citet{DBLP:conf/ipmi/SchleglSWSL17}) is GAN-based method
for anomaly detection. This method involves training a DCGAN (\citet{DBLP:journals/corr/RadfordMC15})
and using it to recover a latent representation for each test data
sample at inference time. The anomaly score is a combination of reconstruction
and discrimination components.

\subsection{Adversarially Learned Anomaly Detection (ALAD)}

ALAD (\citet{DBLP:conf/icdm/ZenatiRFLC18}) is based on a bidirectional
GAN anomaly detector, which uses reconstruction errors from adversarially
learned features to determine if a data sample is anomalous. ALAD
employs spectral normalization and additional discriminators to improve
the encoder and stabilize GAN training.

\subsection{Deep Support Vector Data Description (DSVDD)}

DSVDD (\citet{DBLP:conf/icml/RuffGDSVBMK18}) trains a deep neural
network while optimizing a data-enclosing hypersphere in the output
space.

\subsection{One-Class Support Vector Machine (OC-SVM)}

The OC-SVM (\citet{DBLP:conf/nips/ScholkopfWSSP99}) is kernel-based
method that learns a decision boundary around normal examples. 

\subsection{Isolation Forest (IF)}

An isolation forest (\citet{DBLP:conf/icdm/LiuTZ08}) is a partition-based
method which isolates anomalies by building trees using randomly selected
split values.

\section{Commonly Used Datasets in The Cyber Security Domain}

One of the major issues in cyber security related research is that,
unlike the computer vision domain, there are very few publicly available
datasets containing malicious samples, as such samples are either
private and sensitive (e.g., benign emails for spam filters) or can
cause damage (e.g., PE malware samples). Moreover, the cyber security
domain is rapidly evolving, so older datasets do not always accurately
represent the relevant threats in the wild. The lack of common datasets
makes the comparison of attacks and defenses in the cyber security
domain challenging.

Table \ref{tab:Commonly-Used-Datasets} lists some of the datasets
that are publicly available in the cyber security domain and commonly
used to train the attacked classifiers mentioned in this paper.

\begin{table}
\caption{\label{tab:Commonly-Used-Datasets}Commonly Used Datasets in the Cyber
Security Domain}

\centering{}%
\begin{tabular}{>{\centering}p{0.08\textwidth}>{\centering}p{0.09\textwidth}>{\centering}p{0.04\textwidth}>{\centering}p{0.1\textwidth}>{\centering}p{0.2\textwidth}>{\centering}p{0.4\textwidth}}
Dataset's Name & Citation & Year & Input Type & Number of Samples & Extracted Features/Raw Data\tabularnewline
\midrule
DREBIN & \citet{DBLP:conf/ndss/ArpSHGR14} & 2014 & APK Malware & 5,600 & Permissions, API calls, URL requests\tabularnewline
\midrule
EMBER & \citet{DBLP:journals/corr/abs-1804-04637} & 2018 & PE Malware & 2,000,000 & PE structural features (byte histogram, byte entropy histogram, string
related features, COFF header features, section features, imports
features, exports features)\tabularnewline
\midrule
NSL-KDD & \citet{Tavallaee:2009:DAK:1736481.1736489} & 2009 & Network Stream (PCAP) & 4,900,000 & Individual TCP connection features (e.g., the protocol type), domain
knowledge-based features (e.g., a root shell was obtained) and statistical
features of the network sessions (e.g., the percentage of connections
that have SYN errors in a time window)\tabularnewline
\midrule
CSE-CIC-IDS2018 & \citet{Sharafaldin18} & 2018 & Network Stream (PCAP) & The attacking infrastructure includes 50 machines, and the victim
organization has 5 departments and includes 420 machines and 30 servers & The dataset includes the captured network traffic and system logs
of each machine, along with 80 statistical network traffic features,
such as the duration, number of packets, number of bytes, length of
packets\tabularnewline
\midrule
BoT-IoT  & \citet{DBLP:journals/corr/abs-1811-00701} & 2018 & IoT Network Stream (PCAP) & 72,000,000 & Raw data (PCAPs)\tabularnewline
\midrule
Trec07p & \citet{kuleshov2018adversarial}%
{} & 2007 & Spam Emails (HTML) & 50,200 spam emails and 25,200 ham (non-spam) emails & Raw data (HTML)\tabularnewline
\midrule
VGG Face & \citet{Parkhi15} & 2015 & Images and corresponding face detections & 2,600 & Raw data (images)\tabularnewline
\midrule
IEMOCAP & \citet{DBLP:journals/lre/BussoBLKMKCLN08} & 2008 & Speech (audio) & 10 actors: 5 male and 5 female; 12 hours of audio & Raw waveforms (audio)\tabularnewline
\bottomrule
\end{tabular}
\end{table}

\end{document}